\documentclass{article}

\usepackage{latexsym}

    \usepackage[final]{neurips_2022}

\usepackage{mathtools} 
\usepackage{booktabs} 
\usepackage{tikz} 

\usepackage{float} 

\usepackage{times}
\usepackage{subfigure}

\usepackage{algorithm}
\usepackage{algorithmic}

\usepackage{sidecap}

\usepackage{hyperref}

\usepackage{amssymb} 

\renewcommand\algorithmiccomment[1]{
  {
  	{
  	{\textit{\%\ #1}}
  	}
  }
}

\definecolor{shadecolor}{gray}{0.95}

\usepackage{xcolor}         
\usepackage[normalem]{ulem}

\title{JAF: Judge Agent Forest}

\author{
  Sahil Garg, Brad Cheezum, Sridhar Dutta, Vishal Agarwal \\
  AI Research at Averlon\\
  Redmond, WA\\
  Correspondence: \texttt{sahil.garg@averlon.io, sahil.garg.cs@gmail.com} \\
}

\begin{document}

\maketitle

\begin{abstract}
Judge agents are fundamental to agentic AI frameworks: they provide automated evaluation, enforce domain-specific guardrails, and enable iterative self-refinement of reasoning processes. We introduce \emph{JAF: Judge Agent Forest}, a framework in which the judge agent conducts \emph{joint} inference across a \emph{cohort} of query--response pairs generated by a primary agent, rather than evaluating each in isolation. Here, a cohort denotes a logically related collection of instances (for example, all open vulnerability-issues for a tenant, account, or time window), not a mini-batch chosen purely for computational efficiency. This paradigm elevates the judge from a local evaluator to a holistic learner: by simultaneously assessing related responses, the judge discerns cross-instance patterns and inconsistencies, whose aggregate feedback enables the primary agent to improve by viewing its own outputs through the judge's collective perspective. By situating joint inference at the judge layer, JAF maintains agent modularity while promoting information flow across inputs, fostering consistency, and allowing individual instances to benefit from shared context during evaluation.

Conceptually, JAF bridges belief propagation and ensemble-learning principles: overlapping in-context neighborhoods induce a \emph{knowledge-graph} structure that facilitates propagation of critique, and repeated, randomized evaluations yield a robust ensemble of context-sensitive judgments. JAF can be instantiated entirely via in-context learning (ICL), with the judge prompted for each query using its associated primary-agent response plus a small, possibly noisy set of peer exemplars. While $k$-nearest-neighbor selection in embedding space is a natural starting point for exemplars, this approach often overlooks categorical structure, domain metadata, or nuanced distinctions accessible to modern LLMs.

To overcome these limitations, we develop a flexible locality-sensitive hashing (LSH) algorithm that learns informative binary codes by integrating information-theoretic principles, semantic embeddings, LLM-driven hash predicates, supervision from categorical labels, and relevant side information. These hash codes support efficient, interpretable, and relation-aware selection of diverse exemplars, and further optimize test-time compute by guiding exploration of varied chain-of-thought reasoning paths; optional reinforcement learning can bias this exploration toward underrepresented hash regions. We validate JAF with an empirical study on the demanding task of vulnerability triage in large-scale cloud environments.

\end{abstract}
    
\section{Motivating Use Case: Triaging Vulnerabilities in Cloud Environments}
\label{sec:triage}
Vulnerability triage in cloud environments is the primary application setting for this work and serves as the running example that grounds our framework. This domain exemplifies multi-step, context-dependent reasoning spanning software inventories, network topology, identity and access policies, and configuration state, all of which must be integrated in the face of incomplete and noisy signals to produce a coherent risk posture. The resulting decisions hinge on subtle interactions among these factors rather than any single signal, making the task representative of the challenges faced by realistic agentic systems. While Section~\ref{sec:triage} provides critical motivation for the specific challenges addressed by Judge Agent Forest (JAF), readers interested primarily in general machine-learning aspects may proceed directly to Section~\ref{sec:intro_jaf}.

Modern cloud-scale environments are confronted with an unprecedented volume of vulnerability disclosures across virtual machines, containers, serverless functions, and managed services. Distilling this stream of findings into prioritized, actionable decisions—such as which risks to remediate urgently, monitor, or accept—demands far more than interpreting a CVE identifier or consulting a CVSS score~\citep{mell2007cvss,nist2022cvss}. Effective triage thus requires synthesizing diverse and often conflicting signals: intrinsic vulnerability properties, asset criticality, software roles, network exposure, IAM boundaries, observed attack activity, and operational remediation constraints. Crucially, the real organizational impact and exploitability depend not on any predetermined classification, but on how these dimensions interact within specific deployments.
    
Moreover, CVSS inputs themselves are frequently noisy and inconsistent across sources~\citep{spring2021time,sciencedirect2015cvss,sei2018cvss,arxiv2025cvssempirical}. For the same CVE, public feeds such as NVD, GHSA, and vendor advisories often provide divergent assessments for core metrics (attack vector, complexity, privileges, user interaction). While advisories offer a summary and CVSS vector, the rationale behind each factor is rarely made explicit. Upstream sources like NVD and GHSA adopt broad, minimal-hardening threat models, while vendor advisories may assume non-default security features routinely disabled by users. As a result, conflicting CVSS vectors often reflect differences in environmental assumptions rather than factual disagreements—subtleties almost never surfaced to the triage workflow. Both human and AI-driven triage must treat these vectors not as authoritative, but as provisional signals demanding critical interpretation and contextualization.

\begin{figure}[tp!]
\centering
\subfigure[Isolated judge reviews (standard pattern).]{%
\includegraphics[width=0.45\linewidth]{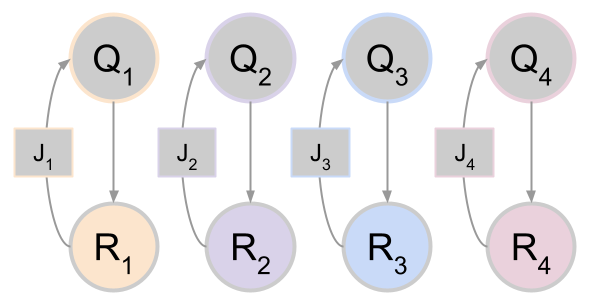}
\label{fig:jaf_isolated}
}
\hfill
\subfigure[Joint judge reviews under JAF.]{%
\includegraphics[width=0.45\linewidth]{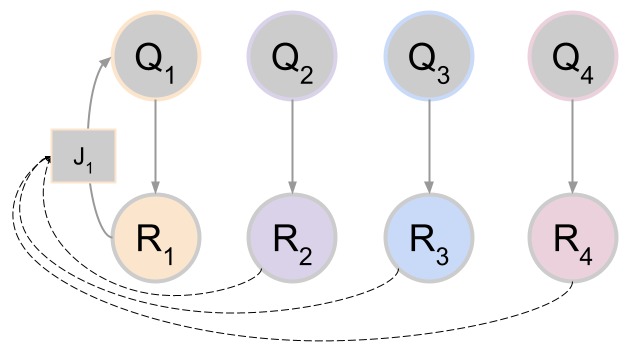}
\label{fig:jaf_joint}
}
\caption{
High-level contrast between standard instance-local use of a judge agent and the joint, cohort-level perspective adopted in Judge Agent Forest (JAF).
In panel~(a), each query $Q_i$ and primary-agent response $R_i$ is evaluated by a judge in isolation; information does not flow across instances.
In panel~(b), a single judge agent reviews each $(Q_i,R_i)$ while also seeing a small set of other query--response pairs from the same tenant- or account-scoped cohort in its prompt.
This richer context allows the judge to compare decisions across related issues, detect inconsistencies, and propagate useful critique patterns, without changing the underlying LLM architecture.}
\label{fig:jaf_isolated_joint}
\end{figure}

Even at the software level, triage is structurally complex. A decision may hinge on the \emph{functional role} of the affected package (e.g., client, server, or library) and whether any associated ports are exposed to untrusted networks. For example, a vulnerability in an interactive client application---such as an email client or terminal shell---may be irrelevant in a headless, automated cloud deployment, even if the client is present, while a flaw in a network-facing server or daemon that accepts unsolicited input constitutes a critical risk. Many packages do not fit a single category; deployments, configurations, and documentation further blur boundaries. Similarly, practical risk assessment hinges on mapping expected ports to those actually reachable in the deployment, accounting for helper modules, dynamic policies, and segmentation. Overlooking auxiliary ports can hide exposures, while ignoring restrictive network controls can overstate risk. Thus, robust triage requires context-aware inference over both software role and port exposure—seemingly basic properties that, in practice, require nuanced and adaptable reasoning.

Cloud-native deployments further complicate this picture~\citep{pohl2021continuous}. The same vulnerable package can pose drastically different risks depending on its location in the infrastructure: network topology (VPC layouts, security groups), effective reachability, IAM policies~\citep{dod2024iam,arxiv2024iamperception,csa2025agenticiam}, and the presence or absence of hardening. Misconfigurations---such as permissive rules or exposed management endpoints---dramatically amplify the impact of issues that look modest in generic metadata~\citep{owasp2021misconfig}. Container orchestration and Kubernetes add further failure modes, including unsafe base images, control plane exposures, and cluster misconfigurations that broaden attack surfaces~\citep{kubegrade2025security,sentinelone2024kuberisks,aikido2025kube}.

Robust triage is thus a multi-stage reasoning problem over fragmented and heterogeneous evidence, requiring continual synthesis across software, configuration, and observed state. Initiatives such as SBOMs~\citep{ntia2021sbom,wm2025sbom,arxiv2024sbomlandscape,sciencedirect2025sbom} help, but do not remove the need for high-level contextual aggregation and inference.
        
\begin{figure}[tp!]
\centering
\includegraphics[width=0.85\linewidth]{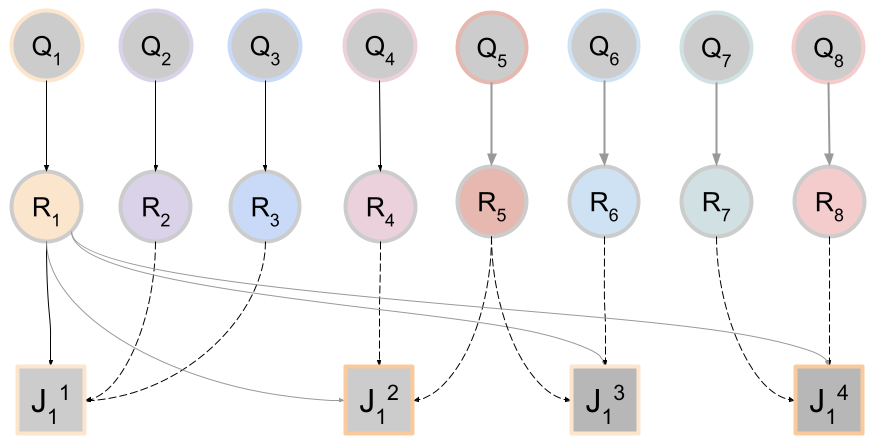}
\caption{
Conceptual overview of JAF operating on a cohort of queries.
Nodes $Q_1,\dots,Q_8$ (top row) represent queries (e.g., vulnerability--asset pairs), and nodes $R_1,\dots,R_6$ (middle row) represent the corresponding primary-agent responses.
Each judge invocation $J^1,J^2,J^3,J^4$ (bottom row) reviews one focal response while also receiving a random, relation-aware subset of other query--response pairs, indicated by the dashed connections.
Across the cohort, these overlapping neighborhoods induce a ``forest'' of judge contexts: each $J^k$ is a tree rooted at a focal instance, with branches to its in-context exemplars.
Repeated passes over the same cohort (or over successive cohorts from the same tenant or account) (or multiple evaluation runs) allow information to propagate along these connections, leading to more calibrated and consistent judgments than isolated, one-shot reviews.}
\label{fig:jaf_batch_forest}
\end{figure}

From an AI perspective, cloud vulnerability triage is compelling because it demands the full arsenal of agentic reasoning: long-horizon, tool-augmented inference over structured and unstructured data; robust calibration of decisions under uncertainty (balancing false negatives and false positives); and inherently \emph{cohort-based} decision-making, where sets of related vulnerabilities must be prioritized collectively for a tenant, time window, or campaign. Prioritization requires integrating CVSS metrics, external signals such as EPSS~\citep{ssrn2020epss,acm2021epss,blackhat2019epss}, and organizational asset context. Cohorts are typically scoped to a single tenant or cloud account, enabling rich context reuse while ensuring privacy and separation.

These properties make triage an ideal proving ground for next-generation agentic AI and LLM-based judge frameworks. The central challenge: many meaningful signals and useful regularities only become accessible when decisions are considered jointly across sets of related inputs. Instance-local judgment fails to capture these dependencies and emergent cohort-level patterns. The ideas and methods we develop here, though grounded in cloud vulnerability triage, are broadly applicable to any context where large cohorts of model decisions require coherent, context-aware refinement.

\section{Introduction to JAF}
\label{sec:intro_jaf}

Judge agents have emerged as a key component in agentic AI systems~\citep{moveworks2026agentic,exabeam2025agentic,akka2025agentic,shinn2023reflexion,madaan2023self,zheng2023judging,yao2023react,schick2023toolformer,zelikman2022star}: they evaluate model outputs, provide guardrails in complex workflows~\citep{arxiv2025sguard,arxiv2024safeguarding}, and support iterative self-refinement of reasoning~\citep{madaan2023self,shinn2023reflexion}. In a vulnerability triage setting, for example, a primary agent might propose that a given issue is "medium risk" because it affects an internal-only service with network segmentation and read-only file systems in place; a judge LLM can then verify that this claim is actually supported by the collected evidence, that no conflicting signals (e.g., exposed management ports or overly permissive IAM roles) have been overlooked, and that the reasoning aligns with organizational policy. Depending on where it is inserted in the pipeline, the same judge mechanism can act as a fact-checker for environment discovery, a consistency checker for mitigation inference, a calibrator for risk scoring, or a final gatekeeper that flags hallucinations, missing evidence, or unsafe recommendations before they reach human analysts or downstream automation.
    
Most current agentic workflows, including those that employ judges, operate in a fundamentally \emph{instance-local} fashion~\citep{zelikman2022star}.
As schematized in Figure~\ref{fig:jaf_isolated_joint}(a), each query $Q_i$ is processed end-to-end in isolation, and any associated judge sees only its corresponding primary-agent response $R_i$.
In-context examples, when used at all, typically come from static human annotations or previously labeled instances~\citep{dong2023survey}. 
But---to the best of our knowledge---no prior work (except for the very recent work by \cite{bot2026} which has been accomplished parallel to ours) treats the \emph{current tenant- or account-level cohort} of inputs as a coupled object of inference rather than a set of independent cases (batching for compute efficiency is a separate concern that is a standard trick complementary to our work).
In vulnerability triage, this means that even though a tenant views all of their outstanding issues together on a single dashboard, the agentic pipeline is usually unaware of the joint impact, relative risk, and recurring environmental quirks that span those issues.
A purely instance-local judge cannot systematically exploit patterns such as "all services behind this unusual network segment behave similarly," enforce consistency in how similar assets or vulnerabilities are classified, or propagate a hard-won insight: if a subtle misconfiguration is recognized for one ticket, there is no principled way to reflect that discovery back onto related tickets.
These limitations point to a missing capability: a mechanism for the judge to perform genuinely \emph{joint}, cohort-level reasoning over related query--response pairs, so that cohort-wide regularities, dependencies, and rare-but-crucial signals can be captured and reused, while the underlying task agents remain modular and instance-wise.

\begin{figure}[tp!]
  \centering
  \subfigure[Iterative self-refinement for a single query.]{%
    \includegraphics[width=0.15\linewidth]{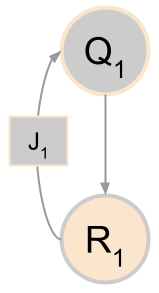}
    \label{fig:jaf_self_refine}
  }
  \hfill
  \subfigure[Multiple chains of thought (CoT) for a single query.]{%
    \includegraphics[width=0.3\linewidth]{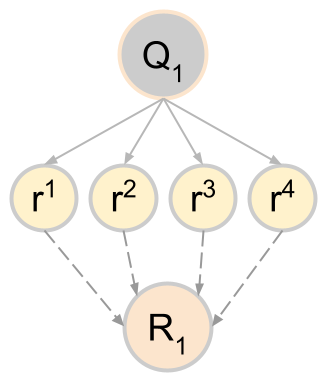}
    \label{fig:jaf_cots}
  }
  \hfill
  \subfigure[JAF unifies self-refinement and CoT exploration across related queries.]{%
    \includegraphics[width=0.45\linewidth]{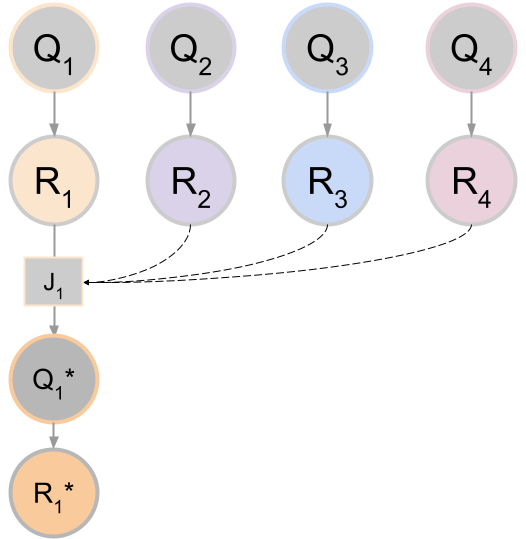}
    \label{fig:jaf_joint_cot}
  }
  \caption{
  How Judge Agent Forest (JAF) combines iterative self-refinement and exploration of multiple chains of thought.
  Panel~(a) illustrates a conventional self-critique loop for a single query $Q_1$, where a judge reviews a primary-agent response, produces feedback, and the primary agent updates its answer~\citep{madaan2023self,shinn2023reflexion}.
  Panel~(b) depicts standard CoT exploration, where several independent reasoning paths ($r^1,r^2,\dots$) are generated and later aggregated~\citep{wei2022cot,wang2022selfconsistency}.
  Panel~(c) shows JAF's perspective: the judge sees multiple CoT variants \emph{and} multiple related queries simultaneously, enabling it to compare reasoning patterns across both dimensions.
  This joint view allows the judge to highlight inconsistencies, reuse successful reasoning templates, and guide refinement in a way that is sensitive to the overall distribution of problem instances within a tenant or account cohort, not just a single case.}
  \label{fig:jaf_cot_refine}
\end{figure}
    
Building on the need for joint evaluation, we investigate: \emph{how does the structure and capability of agentic workflows change when a judge reasons over related query--response pairs in concert, rather than treating each in isolation?} In such a regime, when the judge reviews a particular input--output pair, its prompt includes not only the target instance but also a small, intelligently-chosen set of other query--response pairs from the same cohort. Information thus flows across these local neighborhoods within a cohort: an insight about a misconfiguration or previously unseen mitigation found in one case (such as an overlooked firewall rule) can directly inform the judgment of related instances in the same tenant's triage cohort. Importantly, this mechanism enables not only the primary agent but also downstream consumers of its output to iteratively refine or recalibrate their reasoning as corrections and insights discovered for one cohort member propagate to others, providing a feedback loop that sharpens global understanding and decision consistency throughout the agentic workflow. For example, if a triage system uncovers non-obvious port exposure for a particular service, this knowledge can retrospectively influence risk assessment or prioritization for other seemingly-independent vulnerabilities within the same organizational context.

\begin{figure}[tp!]
\centering
\includegraphics[width=0.9\linewidth]{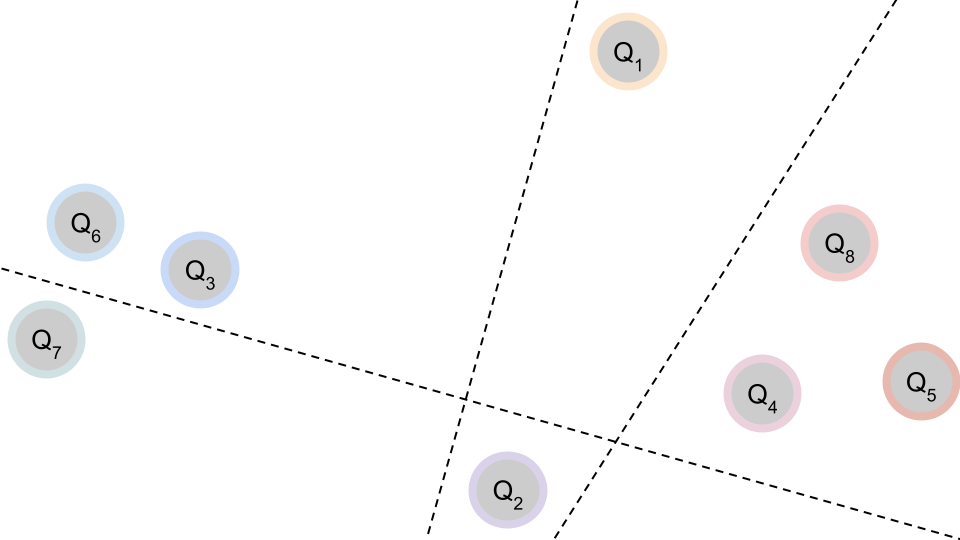}
\label{fig:jaf_lsh_space}
\caption{
A sketch of locality-sensitive hashing (LSH) applied to joint representations of query--response pairs~\citep{indyk1998approximate,andoni2015practical,garg2019kernelized,garg2020modeling}.
Each point corresponds to a pair $(Q_i,R_i)$ embedded in a semantic space that reflects both the input (e.g., vulnerability context, asset metadata) and salient features of the primary-agent response (e.g., cited evidence, inferred exposure).
Dashed lines represent randomized hash hyperplanes; together they define binary hash codes (buckets) such that nearby points are likely to share the same code.
Within JAF, these buckets are used to efficiently select informative and diverse in-context exemplars for the judge:
for a focal pair $(Q_i,R_i)$, we sample other pairs from its own and neighboring buckets to populate the judge's prompt.
This organizes the cohort into a structure that is both computationally tractable and aligned with the underlying semantics and domain structure of the task.}
\end{figure}
        
Formally, the inclusion of additional query--response pairs as context in each judge's prompt induces a dynamic neighborhood structure: each cohort can be viewed as a knowledge graph, where nodes correspond to input--output pairs and edges indicate the co-occurrence of instances within a judge's prompt (see also Figure~\ref{fig:jaf_kg} for a schematic view of this induced graph). Information and critique therefore propagate directly along these local neighborhoods within a single evaluation pass (Zhu and Ghahramani, 2002; Pearl, 1982; Iscen et al., 2019a,b; Garg et al., 2023b). When this cohort-level judging process is run iteratively---for example, in self-refinement regimes where the judge repeatedly critiques and updates outputs---these local communications are compounded: updates begin to percolate through the knowledge graph, gradually allowing patterns or corrections discovered in one neighborhood to affect more distant regions of the cohort. This mechanism effectively implements belief or label propagation (Zhu and Ghahramani, 2002; Pearl, 1982; Garg et al., 2023b), in which feedback provided locally eventually enforces global consistency and allows hard-to-detect signals to reach all relevant instances (see Figure~\ref{fig:jaf_bp} for an abstract illustration of this process within JAF).
        
The way these neighborhoods are constructed adds a further dimension of robustness: by choosing which neighbors to include for each prompt randomly---and by allowing the notion of neighborhood itself to be flexible, potentially defined by semantic similarity, graph distance, metadata, or learned criteria---the system ensures that each input--output pair is reviewed multiple times from diverse contextual perspectives. This randomness is analogous to the construction of classical random forests~\citep{breiman2001random}, where each "tree" (here, each judge pass) sees a different subset of the data through bootstrap aggregating; for a given vulnerability assessment, the judge might draw different neighboring issues to use as context for each trial, thus surfacing different comparative insights or failure modes. Even in sequential self-refinement, where updates naturally propagate through the cohort over repeated iterations, this randomized neighborhood structure means that each instance participates in a diverse set of local contexts over time, functionally mirroring an ensemble effect: conclusions formed in one context can be challenged or reinforced in another, implicitly aggregating evidence in a probabilistic and robust manner. At evaluation time, this approach supports repeated trials with varying context---analogous to ensembling in classical ML evaluation---thereby yielding both more reliable and more nuanced assessments of challenging or ambiguous cases. In the triage domain, this amounts to repeatedly probing the same risk scenario under the lens of multiple environment-specific or vulnerability-specific reference sets, surfacing the true risk profile through consensus over diverse contexts.

\begin{figure}[tp!]
  \centering
  \includegraphics[width=0.85\linewidth]{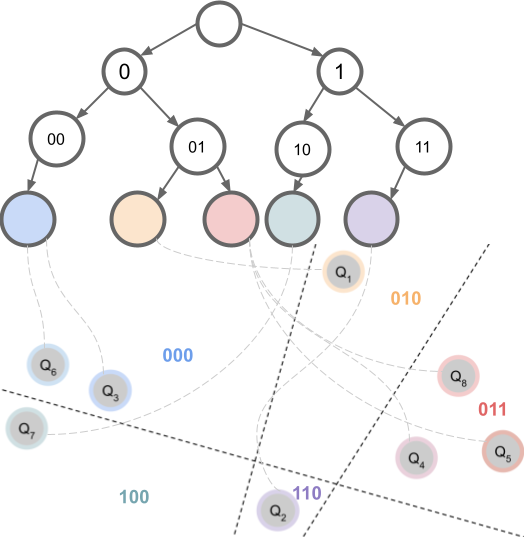}
    \label{fig:jaf_lsh_tree}
  \caption{
  Conceptual view of how LSH can be used to guide exploration of diverse chains of thought (CoTs) within JAF~\citep{openreview2025scalingtest,arxiv2025artscaling,aclanthology2025s1,garg2023clustering}.
  Each node in the tree represents a partial or complete reasoning path for a query; these paths are embedded and hashed into binary codes (e.g., 000, 001, \dots, 111) according to their semantic and procedural characteristics.
  Buckets with many CoTs correspond to frequently explored reasoning styles, while sparsely populated buckets capture under-explored or atypical strategies.
  By monitoring which hash codes are under-sampled and how they correlate with judged quality, JAF can allocate additional test-time compute to promising yet less-visited regions of the reasoning space, optionally using reinforcement learning~\citep{arxiv2024exploration,stanford2018scalable} to learn exploration policies.
  This allows the system to balance exploitation of known-good reasoning templates with exploration of novel ones, all within a principled, LSH-organized framework that is aligned with the emerging knowledge graph over queries and CoTs.}
\end{figure}
        
Together, these intertwined mechanisms---localized information flow, iterative propagation, and ensemble-style contextual diversity---enable a simple yet powerful framework for information flow between agentic workflows operating on instances in a cohort. Within the vulnerability triage domain, this means our approach can systematically exploit both obvious and subtle relationships, such as repeated misconfigurations, rare-but-critical vulnerabilities, or shared network exposure, while maintaining the modularity of each agent. More broadly, our method opens the door to integrating powerful nonparametric techniques---including inference on knowledge graphs and ensemble decision strategies---with the expressive, parametric backbone of LLM-based agents, achieving robust and context-sensitive joint inference in any domain where cohort-level evaluation and decision consistency matter.    
    
Within this joint-evaluation picture, JAF also subsumes prior uses of judges for \emph{iterative self-refinement}~\citep{madaan2023self,shinn2023reflexion} and \emph{multi-chain-of-thought (CoT) exploration}~\citep{wei2022cot,wang2022selfconsistency} on single instances~\citep{yao2023react,schick2023toolformer}.
As illustrated in Figure~\ref{fig:jaf_cot_refine}(a)--(b), a judge can critique and refine a single reasoning chain, or compare multiple independently generated CoTs for the same query.
In JAF, the same underlying mechanism naturally yields these benefits at the cohort level: as shown in Figure~\ref{fig:jaf_cot_refine}(c), the judge can see multiple CoT variants \emph{and} multiple related queries in a single prompt, allowing it to compare, transfer, and refine reasoning patterns across both dimensions.
Rather than explicitly generating many CoTs for each hard query in isolation, the system can exploit the diversity of reasoning that already arises across the cohort, so that CoTs produced for some instances become informative context for others.
This provides many of the robustness and calibration advantages of multi-CoT and self-refinement, while amortizing their compute cost over an entire workload instead of duplicating it per instance.
        
In our formulation, the local ``neighborhoods'' that drive information flow and ensemble behavior are instantiated as the small sets of additional query--response pairs that accompany $(Q_i,R_i)$ in the judge's prompt.
Because the judge's context window is finite, these neighborhoods cannot simply be the entire cohort; they must be selected.
We therefore require a mechanism that, for each query, chooses a compact set of exemplars that are both \emph{relevant} to the focal instance and collectively \emph{diverse} across the cohort.
A natural starting point is to embed queries (and optionally responses) into a semantic vector space and select $k$ nearest neighbors (kNN)~\citep{sciencedirect2020knn,arxiv2025semantic} for each focal query.
This baseline is simple and often effective, but it instantiates neighborhoods purely through embedding geometry and ignores additional structure from labels, historical judgments, domain metadata, or richer, agent-defined features of the responses—limitations that motivate the more flexible LSH-based neighborhoods we introduce next.
        
To this end, we introduce a locality-sensitive hashing (LSH) scheme~\citep{indyk1998approximate,andoni2015practical,garg2019kernelized} over the semantic representations used to form neighborhoods.
Classical LSH uses randomized hash functions (such as random hyperplanes) so that nearby points in an embedding space collide in the same bucket with high probability.
We build on this idea but go beyond purely random projections: we learn a family of hash functions via small neural networks that operate on the (comparatively low-dimensional) semantic embeddings produced upstream.
These networks are optimized using information-theoretic criteria~\citep{tishby2000infobot,arxiv2015deeplearningIB,garg2023clustering} so that the resulting hash codes form synergistic, non-redundant partitions of the space—supporting clustering of related instances and explicit detection of out-of-distribution points~\citep{arxiv2018learningconfidence,arxiv2021bayesianood,garg2023ood}.
This helps suppress noise and irrelevant variation in the embeddings while preserving structure that matters for downstream judgment.
At a coarse level, these learned hash functions induce buckets that capture major semantic regularities across the cohort.
However, JAF occasionally requires finer-grained distinctions within a coarse bucket—for example, when reinforcement learning is used to explore diverse chains of thought in a focused region of the space.
For such cases, we refine the hashing locally by invoking LLM-based agents as hash designers over small subsets: given a handful of instances in a bucket, an agent can either (i) select a reference subset to define an in-distribution region and train an information-theoretic out-of-distribution detector~\citep{garg2023ood} that splits the bucket, or (ii) directly propose a binary split based on high-level textual criteria, without an explicit OOD stage.
In parallel, categorical input features and labels (such as vulnerability classes, asset types, or primary-agent decision labels) can either be treated as hash bits in their own right (after basic transformations like one-hot encoding) or used as supervised signals to train hash functions in an information-bottleneck style.
Together, these mechanisms yield a multi-scale, LSH-style representation in which coarse buckets capture broad semantic similarity, while finer splits capture subtle, task-specific distinctions—all expressed as compact binary codes that are easy to retrieve from and reason over within JAF.
        
Within a cohort, LSH allows us to quickly organize issues into buckets and to draw, for each judge invocation, an informative yet randomized ``mini-forest'' of other pairs to include as in-context examples.
Sampling remains stochastic even with LSH guidance: we never pick a single deterministic neighborhood, but instead draw different subsets across iterations.
This controlled randomness is what endows JAF with its ensemble-like behavior and robustness to noise~\citep{breiman2001random}.

The same structure also supports more advanced reasoning strategies.
By organizing a cohort into LSH buckets, we obtain a natural decomposition of the space of environments and reasoning behaviors.
For particularly difficult queries, we can allocate extra test-time compute by exploring multiple chains of thought or tool-invocation plans, guided by which buckets are under-sampled or historically error-prone.
Figure~\ref{fig:jaf_lsh_tree} sketches this idea at the level of CoT trees: different reasoning paths are embedded, hashed into buckets, and selectively explored to cover diverse, promising regions of the reasoning space.
Reinforcement learning can be layered on top to further bias the system toward CoT paths that historically led to high-quality judged outcomes.
    
Conceptually, JAF is simple: a single judge LLM is repeatedly invoked on each query--response pair together with different, relation-aware random subsets of other pairs from the same cohort.
Yet this simple construction has rich implications for evaluation, self-refinement, and advanced reasoning in agentic systems.
Sections that follow (i) formalize JAF and its connections to classical non-parametric methods, (ii) describe our LSH algorithms over query--response pairs, and (iii) empirically evaluate JAF on the vulnerability triage task introduced in Section~\ref{sec:triage}.
While triage is our primary running example, the underlying ideas are general and apply to any domain where we must judge and improve large cohorts of model decisions.
    
\section{Related Work}
\label{sec:related_work}

Closest in spirit to JAF is the recently proposed Batch-of-Thought (BoT) framework~\citep{bot2026}, which likewise argues that jointly processing related queries enables cross-instance learning that is not available under independent evaluation. BoT was developed concurrently and independently of our work, and is, to our knowledge, the first to systematically study \emph{cohort-level} reflection for LLMs. It is instantiated as cross-instance reasoning in a two-agent ``BoT-R'' architecture: an Actor (a ReAct-style agent~\citep{yao2023react}) produces answer--rationale pairs for a batch of queries, and a Reflector then evaluates all of these pairs jointly in a single shared context, performing comparative analysis to identify inconsistencies, extract shared patterns, and suggest refinements. Although BoT-R allows multiple outer reflection rounds, in each round the Reflector always sees the same fixed batch together; queries are assigned to batches once (sequentially or via semantic clustering) and are never re-batched or exposed under alternative peer groups across iterations. We view this as important early evidence that cross-instance learning at the judge/reflection layer is a promising direction.
    
At the same time, BoT's concrete design choices make it best suited to relatively small, homogeneous batches and moderate-length prompts, whereas JAF is explicitly designed for large, heterogeneous, long-context workloads. In BoT-R, each query belongs to a small, fixed micro-batch (typically of size 4 or 8), and in every reflection round the Reflector receives the \emph{entire} micro-batch context in a single prompt and bases its per-instance refinement flags, confidence scores, and critiques on this global view. This architecture is very effective when inputs are short and structurally similar, but it ties feasibility and performance tightly to batch composition and context limits: in their experiments, the authors report non-monotonic behavior as batch size increases from 1 to 4 to 8, with intermediate batch sizes often yielding the best accuracy--efficiency trade-off, and larger batches suffering from rationale compression (because explanations must be truncated more aggressively to fit into the shared window) and from increased within-batch heterogeneity that weakens comparative signals. They also observe that highly symbolic domains (for example, mathematics and parts of the physical sciences in MMLU~\citep{hendrycks2020mmlu}) can be misled when correlated derivation errors are reinforced by cross-instance agreement, leading to small but consistent drops relative to per-instance reflection, whereas interpretive domains such as humanities~\citep{hendrycks2020mmlu}, social sciences~\citep{hendrycks2020mmlu}, and medicine benefit substantially more from comparative reasoning.

JAF starts from the same high-level insight as BoT---that the judge layer is the natural place to exploit cross-instance structure---but develops it in a way that is tailored to large, heterogeneous, long-context agentic settings such as cloud vulnerability triage. Rather than presenting the judge with an entire tenant- or workload-level cohort at once, JAF operates in a \emph{local, iterative} regime: each judge invocation focuses on a single query--response pair and conditions only on a small, relation-aware neighborhood of other pairs drawn from the same larger pool, with those neighbors represented by compact, task-specific summaries (e.g., primary-agent rationales and key features) rather than full, multi-document contexts. These overlapping neighborhoods induce a sparse knowledge graph over the cohort: two instances are connected if they ever co-occur in a judge's prompt. Successive judge passes with differently sampled neighborhoods then allow critique signals to propagate along this knowledge graph over multiple hops, yielding a soft, language-mediated analogue of graph-based semi-supervision~\citep{zhu2005ssl,blum2001mincut,zhou2004localandglobal} and label or belief propagation~\citep{zhu2002learning,pearl1982belief}. Crucially, the neighborhoods themselves are not fixed by an initial disjoint partition: JAF uses learned, multi-scale locality-sensitive hashing (LSH) that can integrate semantic embeddings, categorical labels, domain metadata, and even LLM-defined hash predicates to organize query--response pairs into interpretable buckets, and then \emph{randomly} samples diverse neighbors from these buckets for each judge call. This LSH-guided, stochastic construction both scales to large cohorts and implements an ensemble-like effect in which each instance is judged multiple times under different, locally coherent contexts---a capability that BoT-R's single global reflector pass over a small, fixed micro-batch does not aim to provide. As a result, JAF remains practical when cohorts contain hundreds or thousands of long, heterogeneous vulnerability--asset cases with rich side information, while still reaping the benefits of cross-instance comparison that BoT demonstrates on smaller, more uniform, and shorter-text benchmarks.
        
Another line of work enhances reliability by aggregating judgments from multiple evaluators rather than multiple instances. CARE~\citep{care2025} casts multi-judge aggregation as inference in a latent-factor Markov random field~\citep{koller2009pgm}, explicitly modeling a latent true-quality variable, inter-judge correlations, and confounders such as length or verbosity. PoLL~\citep{verma2024poll} instead uses a panel of diverse, typically smaller, LLM judges and aggregates their scores via voting or averaging, mitigating intra-model bias and reducing cost relative to a single large evaluator. APE~\citep{ape2025} automatically discovers evaluation dimensions and associated prompts from failure cases, then ensembles judgments across these dimensions using a confidence-aware scheme. All three approaches aggregate \emph{across judges} or \emph{across evaluation criteria} for a fixed instance. JAF is orthogonal: it aggregates information \emph{across related instances} using a \emph{single} judge model. In principle, these axes of diversity can be combined—for example, a CARE- or PoLL-style panel in which each judge itself operates in JAF mode over a cohort, or APE-style dimensions evaluated under JAF's cohort-aware neighborhoods.

A growing body of work also studies how to allocate test-time compute more effectively. \citet{snell2025scaling} show that, for a fixed FLOP budget, scaling test-time search over reasoning paths can be more effective than scaling model parameters, particularly in the "learnable middle range" of task difficulty. OpenAI's o1 models~\citep{openai2024o1} similarly demonstrate that training models to make better use of extended internal chains of thought yields smooth performance gains as both train-time and test-time compute increase. Self-consistency~\citep{wang2022selfconsistency} can be seen as an earlier incarnation of this idea: multiple CoTs are sampled for a \emph{single} query and the final answer is selected by majority vote. All of these methods scale compute \emph{within} an instance, typically by sampling or searching over many trajectories for that input. JAF provides a complementary axis: it amortizes reasoning diversity \emph{across} instances in a cohort. CoTs generated for some queries become informative context for others, and LSH-structured neighborhoods can guide where to allocate extra test-time compute (for example, by exploring additional CoTs in under-explored or historically error-prone regions of hash space). In this sense, JAF's cohort-level structure and induced neighborhood knowledge graph provide a substrate on which internal test-time scaling methods can operate, rather than an alternative to them.
    
\section{Deep Dive into JAF}
\label{sec:deepdive_jaf}
        
Judge Agent Forest (JAF) operationalizes the central idea of this paper: that a judge agent should move from instance-local evaluation to joint, cross-instance judgment over related query--response pairs within a logical cohort of issues (for example, all vulnerability--asset pairs for a tenant, account, or time window), rather than evaluating each decision in isolation. By having the judge LLM review a focal pair $(Q_i, R_i)$ together with a small, task- and context-aware set of other pairs from the same cohort---where relatedness may be defined by semantic proximity, shared vulnerability class, similar assessed severity, common deployment patterns, historical failure modes, or other workload-specific structure---JAF allows individual decisions to be calibrated against their peers, surfacing cross-instance patterns, inconsistencies, and rare-but-critical signals that only emerge when decisions are considered jointly. These overlapping neighborhoods of in-context exemplars induce a knowledge-graph structure over the cohort (Figure~\ref{fig:jaf_kg}), and repeated judging passes with randomized neighborhoods amount to a form of language-mediated, nonparametric propagation of critique and corrections (schematically illustrated in Figure~\ref{fig:jaf_bp}), linking ideas from belief (label) propagation, random forests, and chain-of-thought ensembling. Throughout, we use ``cohort'' in this sense of a logical, semantically coherent collection of issues, not merely as a computational mini-batch.

Let $\mathcal{B} = \{(Q_i, R_i)\}_{i=1}^N$ denote such a cohort of query–response pairs produced by a primary agent. Here $Q_i$ is an input (for example, a vulnerability–asset pair and its associated context) and $R_i$ is the primary agent’s output (for example, a risk assessment with chain-of-thought rationale). A judge agent $J$ is a large language model that, given a focal pair and a set of neighbors from the cohort, returns a discrete judgment and feedback.

In a single JAF pass, for each $i$ we construct a neighborhood $S_i \subseteq \mathcal{B} \setminus \{(Q_i,R_i)\}$ and call the judge on the focal pair plus its neighbors:
\[
(J_i^l, J_i^c) = J\bigl((Q_i,R_i), S_i\bigr),
\]
where $J_i^l \in \{\text{accept},\text{refine}\}$ is a binary label for the focal decision, and $J_i^c$ is a natural-language critique or justification. As in Figure~\ref{fig:jaf_isolated_joint}(b), the prompt to $J$ contains the focal instance and the elements of $S_i$, summarized as necessary to fit the context window. In practice, each $S_i$ is small (dozens of pairs at most), both for context-budget reasons and because we want neighborhoods to remain local and interpretable. Across the cohort, these neighborhoods introduce a dependence structure among instances: the judge’s response for index $i$ depends not only on $(Q_i,R_i)$ but also on the behavior of $(Q_j,R_j)$ for $j \in S_i$.

\subsection{Neighborhoods and the Cohort-Level Knowledge Graph}
        
\begin{figure}[tp!]
  \centering
  \subfigure[JAF neighborhoods over a cohort.]{
    \includegraphics[width=0.75\linewidth]{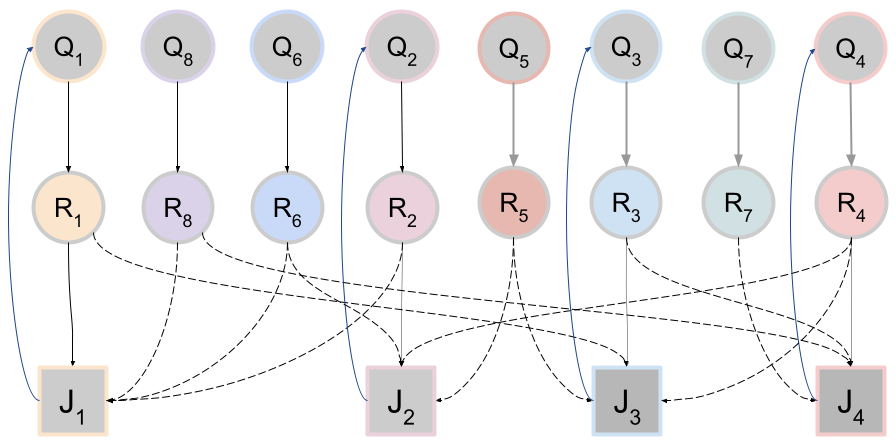}
    \label{fig:jaf_bp_a}
  }

  \vspace{0.5em}

  \subfigure[Initialization of beliefs]{
    \includegraphics[width=0.475\linewidth]{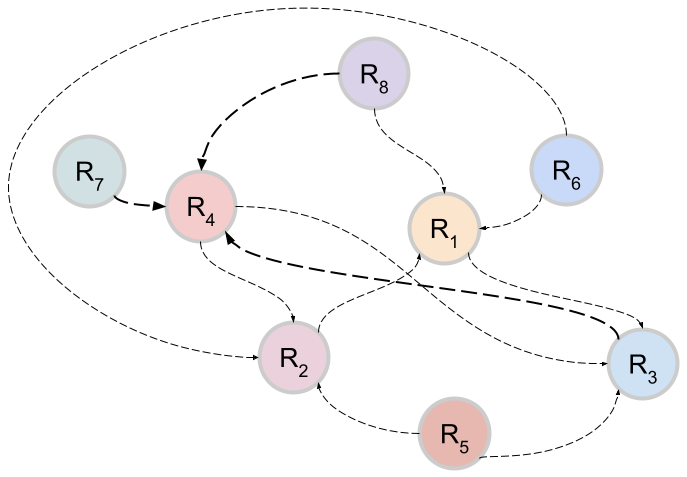}
    \label{fig:jaf_bp_b}
  }
  \hfill
  \subfigure[First round of message passing]{
    \includegraphics[width=0.475\linewidth]{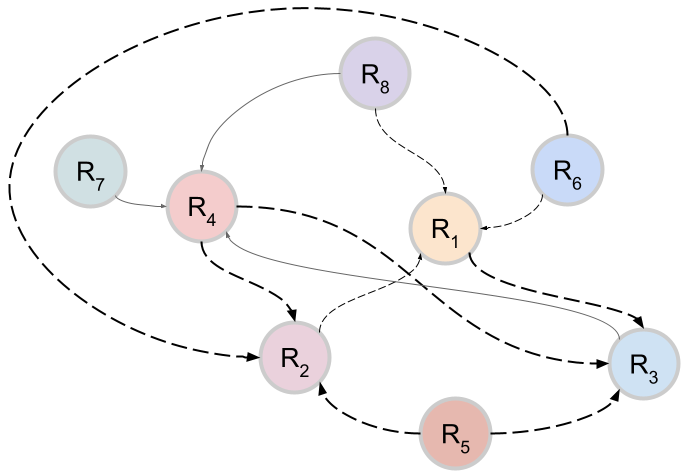}
    \label{fig:jaf_bp_c}
  }

  \vspace{0.5em}

  \subfigure[Subsequent refinement of beliefs]{
    \includegraphics[width=0.475\linewidth]{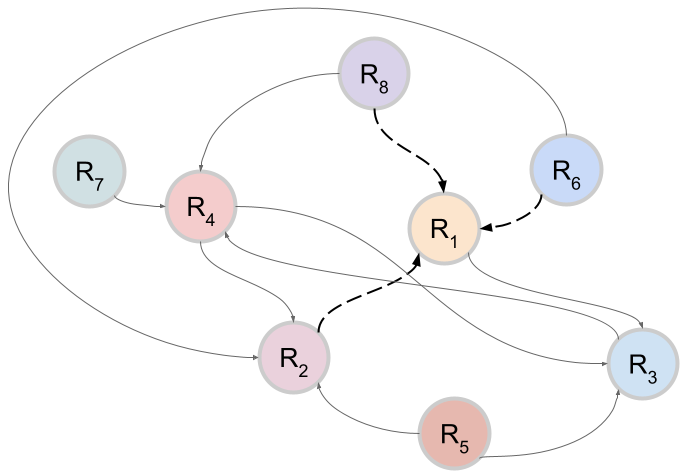}
    \label{fig:jaf_bp_d}
  }
  \hfill
  \subfigure[Converged, globally consistent beliefs]{
    \includegraphics[width=0.475\linewidth]{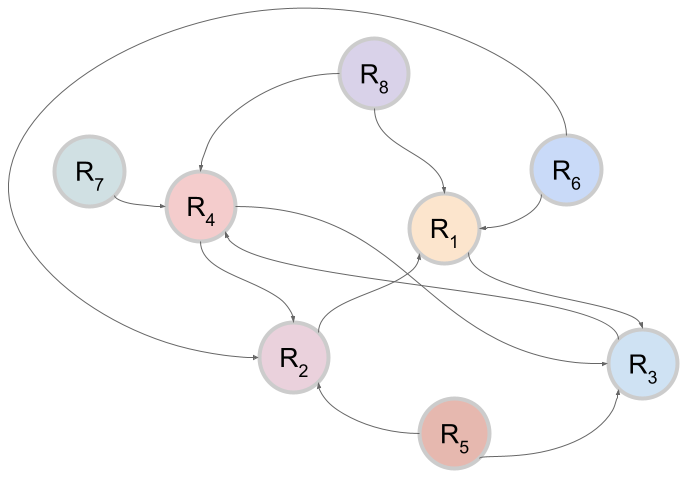}
    \label{fig:jaf_bp_e}
  }

  \caption{
    Relationship between Judge Agent Forest (JAF) and classical belief propagation.
    Panel~(a) illustrates JAF's operating regime over a cohort: each focal instance
    $(Q_i, R_i)$ is judged in the context of a small, overlapping neighborhood of
    other query--response pairs, giving rise to a ``forest'' of judge calls whose
    roots and branches span the cohort.
    Panels~(b)--(e) depict an abstract belief-propagation view of the same structure:
    nodes represent instances, edges connect instances that ever co-occur in a judge
    context, and colors indicate evolving beliefs or confidence scores.
    Starting from locally informed beliefs (panel~(b)), successive rounds of
    message passing (panels~(c) and~(d)) propagate evidence and corrections along
    the edges of this implicit graph, eventually yielding a more globally consistent
    configuration (panel~(e)).
    Running JAF for multiple refinement rounds with re-sampled neighborhoods thus
    approximates loopy belief propagation on the cohort-level knowledge graph, with
    the judge LLM implementing the local update rule at each node.
  }
  \label{fig:jaf_bp}
\end{figure}

\begin{figure}[tp!]
  \centering
  \includegraphics[width=0.65\linewidth]{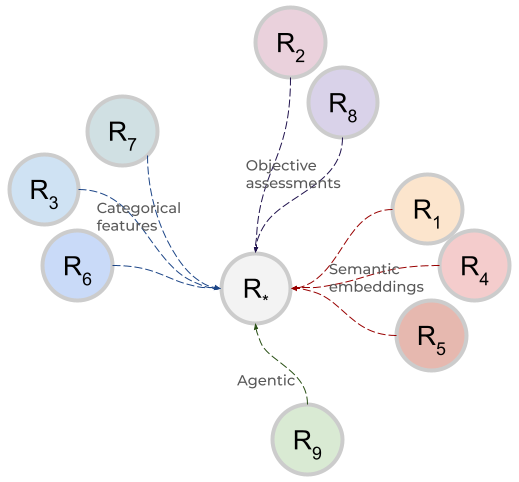}
  \caption{
    Schematic cohort-level knowledge graph used by Judge Agent Forest (JAF).
    Each node corresponds to a query--response pair (for example, a specific
    vulnerability--asset case together with the primary agent's assessment),
    and edges indicate actionable relatedness: two instances are linked if they
    have appeared together in at least one judge neighborhood or if domain
    structure (e.g., shared assets, network segments, or deployment roles)
    ties them together.
    The surrounding colored regions highlight the heterogeneous signals that feed
    into this graph: semantic embeddings, objective assessments (such as risk
    scores or tool outputs), categorical features and metadata (such as CVE family
    or asset type), graph-structural cues from the underlying environment, and
    agentic features derived from chains of thought or tool-invocation traces.
    JAF's neighborhood construction mechanisms (whether kNN-based or LSH-based)
    operate over this knowledge graph to select small, relation-aware sets of
    exemplars for each judge call, enabling critique and corrections to propagate
    along the graph over successive passes.
  }
  \label{fig:jaf_kg}
\end{figure}

We can organize these neighborhoods into a cohort-level knowledge graph $G^{\mathrm{KG}} = (V, E)$. The node set $V = \{1, \ldots, N\}$ indexes instances in the cohort, and an undirected edge $(i, j) \in E$ is present if $(Q_j, R_j)$ appears in $S_i$ or $(Q_i, R_i)$ appears in $S_j$ in at least one judging pass. In this view, the family of neighborhoods $\{S_i\}$ defines a (possibly time-varying) adjacency structure over the cohort; see Figure~\ref{fig:jaf_kg} for an illustration. An edge in $G^{\mathrm{KG}}$ represents actionable relatedness as seen by the judge's context: two issues have been juxtaposed for joint consideration at least once. This does not imply that the underlying queries are nearest neighbors in an embedding space. The situation mirrors classical knowledge graphs, where edges encode heterogeneous, task-defined relations, and continuous embeddings live alongside the graph as auxiliary node features rather than as the source of the topology. Figure~\ref{fig:jaf_bp} complements this static view by visualizing how beliefs or confidence scores can evolve over this graph as JAF runs iteratively.

When JAF is run once on a cohort, the resulting $G^{\text{KG}}$ is sparse and captures which instances were directly compared under at least one notion of relatedness. When JAF is run iteratively with re-sampled neighborhoods, this knowledge graph becomes denser and acquires temporal structure: new edges appear as different neighborhood draws bring previously distant instances together in the judge’s context, and existing edges may be revisited under slightly different summaries or prompts. Over multiple passes, a correction or insight associated with a single pair $(Q_j,R_j)$ can thus influence increasingly distant parts of the cohort, as it is propagated along chains of co-occurrence in $G^{\text{KG}}$.

This knowledge-graph view makes the connection to nonparametric learning explicit. If we interpret $J_i^l$ as a label or belief about instance $i$, successive iterations of JAF can be seen as message passing on $G^{\text{KG}}$, driven by a parametric but black-box update rule (the judge LLM) instead of explicit kernel functions~\citep{zhu2002learning,pearl1982belief}. Critiques $J_i^c$ modify the primary agent’s responses, which in turn change the effective “features’’ $(Q_i,R_i)$ seen by the judge at the next iteration, altering both local decisions and downstream neighborhood composition. In this sense, JAF realizes a language-mediated analogue of belief propagation and label propagation on a cohort-level knowledge graph.

\subsubsection{Explicit Knowledge-Graph Neighborhoods for a Single JAF Pass}

Conceptually, the neighborhoods $S_i$ used by JAF can be obtained by explicitly constructing a binary cohort-level knowledge graph and sampling from its adjacency. The graph $G^{\text{KG}}$ is built from multiple signals. Semantic embeddings provide one source of evidence: pairs $(Q_i,R_i)$ and $(Q_j,R_j)$ that are close in an embedding space can be connected by similarity edges. But additional, task-specific relations are equally important. Two instances may be connected because they share a vulnerability family or asset type; because they refer to the same software component deployed in different network segments; because they have similar assessed severity or mitigation strategies; or because historical data show correlated judge failures or human overrides. Each such relation contributes candidate edges, and after applying simple pruning rules (for example, maximum degree constraints or tenant-level filters) we obtain a sparse, binary adjacency $E$ that encodes whether two instances are deemed actionably related.

In practice, constructing $G^{\text{KG}}$ starts from this pool of candidate relations and then prunes them to obtain a graph that is both informative and tractable. For example, we may add an edge between $i$ and $j$ whenever they share a CVE identifier, belong to the same vulnerability class on similar asset types, or fall within a $k$-nearest-neighbor radius in embedding space, and then drop edges that cross tenant or environment boundaries or that would lead to very high-degree hubs. Domain metadata (such as tenant, environment, or application role) and historical labels (such as prior judge decisions or ground-truth triage outcomes) can guide these pruning choices, but at the end of this stage $G^{\text{KG}}$ is simply a binary, undirected graph over the cohort: an edge $(i,j)$ is either present or absent.

Given such a knowledge graph, a single JAF pass can be described directly in terms of sampling from its neighborhood structure. For each instance $i$, we consider its adjacency list $N(i) = \{j : (i,j) \in E\}$ and build $S_i$ by sampling up to $k$ neighbors from $N(i)$. This preserves the flexibility of the neighborhood definition—any signal that contributes edges to $G^{\text{KG}}$ can affect which peers the judge sees—while keeping per-invocation context size bounded. The corresponding procedure is summarized in Algorithm~\ref{alg:jaf_kg}. For a cohort of size $N$ and neighborhood size $k$, we still make $N$ judge calls per pass; what changes, relative to a purely embedding-based baseline, is that neighbor selection is driven by the binary knowledge graph rather than by a single similarity metric.
    
This explicit construction is not strictly required in deployed systems. In many cases, the knowledge graph remains implicit, with neighborhood sampling implemented by querying indices or services that encapsulate the underlying relations—for example, an approximate nearest-neighbor index augmented with filters over vulnerability classes or asset types, or a retrieval service that returns “related tickets’’ based on historical triage outcomes. Conceptually, however, Algorithm~\ref{alg:jaf_kg} makes clear that JAF’s neighborhoods are defined by the edges of a cohort-level knowledge graph: any form of actionable relatedness that can be encoded as such edges—including, but not limited to, kNN similarity—can be used to populate the judge’s context.

\paragraph{Limitations of Fixed Knowledge-Graph Neighborhoods.}

A binary, hand-constructed knowledge graph provides a clean and flexible way to encode actionable relatedness, but it also has limitations that motivate the richer LSH-based neighborhoods introduced next. First, the quality of $G^{\text{KG}}$ depends on the choice of relation types and pruning heuristics; these are often workload-specific and require manual tuning. Second, a single, fixed graph struggles to capture multi-scale structure: we may want very tight neighborhoods for some decisions and broader, cross-cutting neighborhoods for others, but the binary adjacency does not by itself provide a natural hierarchy of neighborhoods. Third, as workloads evolve over time—new tenants, new vulnerability families, new infrastructure patterns—the handcrafted relations underlying $G^{\text{KG}}$ can become brittle or outdated, especially in regions of the space that are rare or out-of-distribution.

The locality-sensitive hashing (LSH) machinery in the following subsection addresses these issues by learning compact binary codes over query–response pairs (and over chains of thought) that implicitly induce their own neighborhood structure. These learned codes can integrate embeddings, metadata, labels, and LLM-defined predicates into a multi-scale organization of the cohort, yielding neighborhoods that are still binary and graph-like at the level of adjacency, but are now derived from information-theoretic and OOD-aware criteria rather than from a fixed set of hand-engineered relations.

\subsection{Conceptual View of LSH-Based Neighborhoods}

\begin{algorithm}[t]
\caption{Single-Pass JAF with Knowledge-Graph Neighborhoods}
\label{alg:jaf_kg}
\textbf{Require:} Cohort $\mathcal{B} = \{(Q_i,R_i)\}_{i=1}^N$, binary knowledge graph $G^{\text{KG}} = (V,E)$ over indices $\{1,\dots,N\}$, neighborhood size $k$, judge LLM $J$.\\
\textbf{Ensure:} Judge outputs $\{(J_i^l,J_i^c)\}_{i=1}^N$.
\begin{algorithmic}[1]
\FOR{$i = 1$ to $N$}
    \STATE Let $N(i) \gets \{j : (i,j) \in E\}$ be the neighbors of $i$ in $G^{\text{KG}}$.
    \STATE Sample up to $k$ indices $\mathcal{N}_i \subseteq N(i)$ (for example, uniformly at random).
    \STATE $S_i \gets \{(Q_j,R_j) : j \in \mathcal{N}_i\}$
    \STATE $(J_i^l,J_i^c) \gets J\bigl((Q_i,R_i), S_i\bigr)$
\ENDFOR
\end{algorithmic}
\end{algorithm}

\subsubsection{Feature Representation and Hash Buckets}

To overcome the limitations of fixed, hand-constructed knowledge-graph neighborhoods discussed above (and of pure kNN baselines) and to better organize large, heterogeneous logical cohorts, JAF uses a hash-based neighborhood scheme inspired by locality-sensitive hashing (LSH). Conceptually (Figure~\ref{fig:jaf_lsh_space}), for each query--response pair $(Q_i,R_i)$ we first construct a feature representation
\[
x_i = \bigl(x_i^{Q},\,x_i^{R}\bigr) \in \mathcal{X},
\]
where $x_i^{Q}$ aggregates information derived primarily from the input side (for example, textual context, semantic embeddings of $Q_i$, and categorical metadata such as vulnerability class or asset type), and $x_i^{R}$ aggregates information derived from the response side (for example, semantic embeddings of $R_i$, intermediate tool outputs, and objective assessments such as preliminary risk scores or flags produced by the primary agent). Given a set of binary hash functions
\[
h_1, h_2, \dots, h_B : \mathcal{X} \to \{0,1\},
\]
we then map $x_i$ to a binary hashcode
\[
\boldsymbol{c}_i \;=\; \bigl(h_1(x_i),\,h_2(x_i),\,\dots,\,h_B(x_i)\bigr) \in \{0,1\}^B.
\]
    
Instances that the learned hash functions and resulting codes $\boldsymbol{c}_i$ deem \emph{actionably related for judging}---whether because of semantic proximity, shared vulnerability class, similar severity profile, common deployment patterns, recurring misconfigurations, or historical judge behavior---tend to share many bits in their codes (or lie at small Hamming distance), while actionably unrelated instances are separated. These hashcodes define buckets from which we sample neighbors for the judge. 
    
Operationally, JAF uses these buckets directly: for a focal instance we retrieve other instances that share (or nearly share) its code and sample neighbors from that pool. Conceptually, and purely as a mental model that connects back to Section~\ref{sec:deepdive_jaf}, one can equivalently view the family of codes $\{\boldsymbol{c}_i\}$ as inducing an implicit binary knowledge graph in which $i$ and $j$ are adjacent when their codes are identical or lie within a small Hamming radius, but no such graph needs to be explicitly materialized in the implementation. 
    
In the remainder of this subsection we discuss desirable properties of such hashcodes, sketch information-theoretic objectives that encourage these properties, and outline one concrete family of algorithms to learn the underlying hash functions $\{h_b\}_{b=1}^B$.
    
\subsubsection{Side Information and Hashcode Desiderata}

\paragraph{Side information as derived labels.}
To express these desiderata more formally, it is helpful to separate out the structured side information we wish the hashcodes to capture. Let
\[
y_i^{Q} \in \mathcal{Y}^{Q}, \qquad y_i^{R} \in \mathcal{Y}^{R}
\]
denote, respectively, categorical or structured features derived from the input side (for example, vulnerability family, asset type, deployment region) and from the response side (for example, discrete risk classes, mitigation categories, or other objective assessments present in $R_i$). We write
\[
Y_i \;=\; \bigl(y_i^{Q},\,y_i^{R}\bigr)
\]
for the combined ``label'' or metadata vector associated with instance $i$. Although $Y_i$ is deterministically extracted from $x_i$ (for example, by parsing out categorical fields or discretized scores), it is convenient to distinguish
\[
X \in \mathcal{X}
\]
as the random feature vector with empirical distribution induced by the cohort, and
\[
Y = g(X)
\]
as the corresponding collection of derived categorical/label features obtained via a fixed featurization map $g$. This separation lets us talk cleanly about how much information the hashcodes retain about specific structured aspects of the data, even though those aspects are ultimately functions of $X$.

\paragraph{Desiderata for hashcodes.}
At a high level, we would like the hashcodes $\boldsymbol{c}_i$ to satisfy three intertwined properties:-
\textbf{1)}. First, they should be \emph{informative} about relevant side information: taken together, the bits should explain as much as possible about $Y$.
\textbf{2)}. Second, they should be \emph{non-redundant} or \emph{disentangled}: each new bit should contribute information that is not already present in earlier bits.
\textbf{3)}. Third, they should be \emph{interpretable and reliable}: at least some of the early bits should align closely with known, human-understandable features or assessments (for example, separating internal-only services from externally exposed services, or high-risk from low-risk issues), so that the resulting buckets can be inspected and validated.
         
Rather than optimizing three separate losses, we conceptually bundle these desiderata into a single mutual-information objective between the code and the side information. Writing $C = \mathbf{h}(X)$ for the random code vector and using the identity
\begin{align}    
H(C) = \sum_j H(c_j) - TC(C),
\end{align}
where $TC(C)$ denotes the total correlation (multivariate mutual information) among the bits, we can decompose
\begin{align}    
I(Y : C) = H(C) - H(C \mid Y) = \sum_j H(c_j) - TC(C) - H(C \mid Y).
\label{eqn:ith}
\end{align}
Maximizing $I(Y:C)$ therefore simultaneously encourages (i) high marginal entropy for each bit $c_j$, (ii) low redundancy across bits via $TC(C)$, and (iii) class- or label-consistent codes via a small conditional entropy $H(C\mid Y)$.

At first glance, one might think that maximizing $I(Y:C)$ would simply drive the hashcodes to become a near-sufficient statistic for $Y$, ignoring structure in $X$ that is not needed to predict labels. Several aspects of \eqref{eqn:ith} and of our construction mitigate this concern. First, once the dependency between $C$ and $Y$ is largely captured (i.e., $H(Y\mid C)$ and $H(C\mid Y)$ cannot be reduced much further), additional bits are driven primarily by the balance between $\sum_j H(c_j)$ and $TC(C)$: new bits must carry entropy and avoid redundancy to improve the objective. In other words, as the label-explaining capacity of $C$ saturates, extra bits are naturally incentivized to capture residual, label-agnostic structure in $X$, rather than merely replicating $Y$. Second, the mapping from $X$ to $C$ is not arbitrary: in practice it is implemented by constrained models (e.g., small neural networks or simple predicates) that must respect smoothness and capacity limits in the input space. This inductive bias encourages codes that are coherent with the geometry and semantics of $X$, not just with $Y$. Third, it is straightforward to reweight the three terms in \eqref{eqn:ith} if desired—for example, by down-weighting the $-H(C\mid Y)$ term or up-weighting the entropy and total-correlation terms—to further temper any tendency to overfit to labels. Finally, standard regularization techniques and early stopping (as in our prior work~\citep{garg2023clustering,garg2023ood}) provide additional practical safeguards against overfitting $C$ to a finite set of $Y$ values. In aggregate, these factors ensure that while $I(Y:C)$ encourages codes that are label-aware, it does not force them to collapse to ``just the labels,'' and leaves substantial room for capturing richer structure in $X$ that is critical for high-quality neighborhoods in JAF.

\paragraph{Informativeness via mutual information.}
Conceptually, we take $I(Y:C)$ itself as the unifying objective. The predictive view
\[
I(Y:C) = H(Y) - H(Y \mid C)
\]
says that good codes $C$ make the side information $Y$ easier to infer, but our primary interest is not to train a classifier: it is to obtain codes whose induced neighborhoods are aligned with labels and metadata while still reflecting richer, unlabeled structure. The structural view
\[
I(Y:C) = \sum_j H(c_j) - TC(C) - H(C \mid Y)
\]
makes this explicit. The term $\sum_j H(c_j)$ encourages each bit to be active and non-degenerate across the cohort; $-TC(C)$ discourages different bits from encoding the same pattern; and $-H(C \mid Y)$ encourages examples sharing the same $Y$ to occupy relatively tight, coherent regions of code space. In practice, this does not mean collapsing each label $Y=y$ to a single code; the other terms ensure that sufficient intra-class variability remains to support fine-grained, label-conditioned neighborhoods.
    
\paragraph{Non-redundancy and disentanglement.}
From the perspective of the global objective $I(Y:C)$, non-redundancy is captured directly by the total correlation term $TC(C)$. If two bits $c_j$ and $c_{j'}$ always move together, then $TC(C)$ is large and $I(Y:C)$ is correspondingly reduced. Minimizing $TC(C)$ therefore pushes the bits toward a disentangled representation in which each bit contributes complementary structure. In practice, we do not require---and do not even seek---perfect independence across all bits under the global data distribution. Many of the hash predicates we use are \emph{local}: they are only intended to be informative inside a restricted region of the space (for example, within a particular hash bucket or label-conditioned subset), and are effectively constant elsewhere. For such bits, the relevant notion of redundancy is not their contribution to $TC(C)$ under the full distribution, but their contribution to total correlation \emph{within the region where they actually vary}, i.e., under a conditional distribution $p(C \mid \text{region})$. When we bisect an existing hash bucket with a new, local hash function, we conceptually measure redundancy only with respect to variation inside that bucket; outside it the bit is inert and should not be penalized. This viewpoint allows us to accommodate low-entropy or partially redundant bits that are valuable for interpretability (e.g., a bit that essentially indicates ``external vs.\ internal exposure'' within a specific class) or for carving out very specific subregions of code space, without forcing every bit to be globally high-entropy and independent. What matters for JAF is not compressing $C$ as much as possible, but organizing the space into neighborhoods that are both semantically coherent and diverse at the scales where the judge actually operates.
    
\subsubsection{Hierarchical Hash Predicates and Interpretability}

\paragraph{Interpretability of early predicates.}
Interpretability can be encouraged by designing early predicates so that they align with simple, domain-relevant properties of the side information $Y$. In the simplest setting with a small number of salient metadata dimensions, we can make this explicit: for the first few bits we may restrict attention to candidate functions $h_b$ that depend primarily on a single metadata coordinate $y^{Q}$ or $y^{R}$ (such as vulnerability family, asset type, or a discretized risk score), and choose those that achieve high mutual information with that coordinate:
\[
h_1 \approx \arg\max_{h} I\bigl(h(X);\,y^{Q,\text{family}}\bigr), \qquad
h_2 \approx \arg\max_{h} I\bigl(h(X);\,y^{R,\text{risk}}\bigr),
\]
and so on. In practice, however, real workloads involve many interacting label and metadata dimensions, so dedicating an independent bit to each coordinate is neither feasible nor desirable. In these settings, we treat the single-coordinate examples as didactic: they illustrate that early predicates should be explainable in terms of coarse, human-understandable distinctions (for example, ``external-facing production services'' vs.\ everything else) while still respecting the underlying semantic structure encoded in embeddings and other features. Crucially, we aim for \emph{all} bits---not only the first few---to be interpretable: each predicate can be described in natural language, even if it depends on multiple features or subtle patterns in the query--response space. The remaining predicates are then free to capture residual, finer-grained structure in the data, subject to the same informativeness and non-redundancy criteria.

\paragraph{Variable effective code length and tree structure.}
In real workloads, this inevitably leads to a spectrum of predicates: some are high-entropy and active across large portions of the dataset, while others are lower-entropy and only informative within restricted regions of the space (for example, within a particular vulnerability family, deployment pattern, tenant, or even a single hash bucket). We regard these specialized, lower-entropy bits as \emph{local} hash predicates: nearly constant outside their region of applicability, yet crucial for further dissecting neighborhoods inside that region. Consequently, different parts of the space can effectively have different ``usable'' code lengths: some cohorts are well-structured by a shallow sequence of predicates, while others require more bits to reach an equally fine-grained partition. This picture is closer to a decision-tree or hierarchical clustering view than to a flat Cartesian product of all bits: we do not expect all $2^B$ combinations of bit values to be populated or meaningful. Instead, each instance follows a path through a sequence of predicates, and its code can be read as that path. Our objective is not to minimize code length per se, but to obtain hash-induced neighborhoods whose depth and shape are well matched to JAF's cohort-level judging. This hierarchical view also mirrors how humans reason about triage backlogs: broad, obvious distinctions (exposure, environment, business criticality) are applied first, and only then are finer-grained distinctions (specific misconfigurations, niche software roles) brought to bear in the regions where they matter. From a geometric perspective, the resulting structure is reminiscent of trees embedded in a hyperbolic space: regions farther from the ``center'' of typical cases require more predicates to describe, reflecting the greater combinatorial diversity of rare, atypical scenarios.

Rather than relying on fixed random projections, we therefore treat hash functions as a collection of learned predicates that are added sequentially, each one introducing a new split somewhere in this implicit hierarchy. Conceptually, this constructs a forest of decision paths over the cohort: early predicates implement coarse, high-level separations that align with major semantic and risk distinctions, while later predicates provide progressively finer refinements within the resulting branches. Each path from the root to a leaf corresponds to a particular combination of properties (for example, ``internet-exposed production service with high business criticality, using a vulnerable TLS library but with strict network segmentation''), and the corresponding leaf or near-leaf region defines a natural neighborhood for JAF's joint judging. The $I(Y:C)$ objective provides a unifying lens on these predicates: they should collectively maintain sufficient entropy to avoid collapse, avoid redundant splits that merely restate earlier questions, and produce codes that are reasonably consistent within label- or metadata-defined groups.
    
\subsubsection{Information-Theoretic Hash Predicates}

Information-theoretic hash predicates give JAF a way to define binary questions over a rich feature space without committing to any specific model architecture. The only learned object at each node is a scalar scoring function $f:\mathcal{X}\to\mathbb{R}$, which is then thresholded to produce a bit $h(x) = \mathbf{1}\{f(x) \ge c\}$. This $f$ can be parameterized by a shallow MLP, a compact transformer, or a lightweight scalar head on top of an existing LLM or embedding model. As long as $f$ is expressive enough to separate the empirical distributions we care about, the information-theoretic objectives that drive predicate selection—based on divergences between induced clusters—remain unchanged. This decoupling of representation choice from the hash-learning objective is what makes the approach flexible: we can freely swap in better semantic embeddings or metadata encoders without redesigning the divergence criteria.
    
\paragraph{Divergence-based clustering splits.}
One class of hash predicates in JAF is derived from information-theoretic clustering via divergence maximization, following our prior work on KL-based clustering~\citep{garg2023clustering}. Let $X$ denote the random feature vector constructed from a query--response pair $(Q,R)$ (e.g., embeddings plus metadata), and let $\mathcal{R}\subset\mathcal{X}$ be a region of interest in feature space (typically, all points sharing a given code prefix). We view $\mathcal{R}$ as an empirical realization of a ``cluster distribution'' $X_{\mathcal{R}}$. A new binary predicate $h:\mathcal{X}\to\{0,1\}$ partitions $\mathcal{R}$ into two child regions
\[
\mathcal{R}_0 = \{x\in\mathcal{R}: h(x)=0\}, \qquad
\mathcal{R}_1 = \{x\in\mathcal{R}: h(x)=1\},
\]
with associated empirical distributions $X_{\mathcal{R}_0}$ and $X_{\mathcal{R}_1}$. Instead of optimizing purely intra-region criteria such as mutual information between points and cluster labels, we define the quality of a split in terms of how different these two child distributions are, using a symmetric KL-divergence objective
\[
\max_{h} \ \hat{D}\bigl(X_{\mathcal{R}_0}\,\|\,X_{\mathcal{R}_1}\bigr) + \hat{D}\bigl(X_{\mathcal{R}_1}\,\|\,X_{\mathcal{R}_0}\bigr),
\]
where $\hat{D}(\cdot\|\cdot)$ denotes an empirical estimate of KL divergence. As shown in~\citet{garg2023clustering}, this KL objective entails the classical mutual-information criterion $I(X: Y)$ while also explicitly maximizing cross-entropy (and hence minimizing overlap) between cluster distributions, thereby capturing both intra- and inter-cluster behavior in a principled way.

Directly estimating KL divergences between high-dimensional distributions is challenging; instead, we estimate them in their dual Donsker--Varadhan form. For two empirical distributions $X_a$ and $X_b$ supported on $\mathcal{R}_a$ and $\mathcal{R}_b$, respectively, the dual form is
\[
D(X_a\|X_b)
\;=\;
\max_{f \in \mathcal{H}}
\left[
\mathbf{E}_{x\sim X_a} f(x)
-
\log \mathbf{E}_{x'\sim X_b} e^{f(x')}
\right],
\]
where $f:\mathcal{X}\to\mathbb{R}$ ranges over a function class $\mathcal{H}$ (e.g., neural networks) and expectations are replaced in practice by empirical averages. This yields an empirical estimator
\[
\hat{D}_{f}(X_a\|X_b)
=
\frac{1}{|\mathcal{R}_a|}
\sum_{x\in\mathcal{R}_a} f(x)
-
\log\left(
\frac{1}{|\mathcal{R}_b|}
\sum_{x'\in\mathcal{R}_b} e^{f(x')}
\right),
\]
and we maximize over $f\in\mathcal{H}$. The key advantage is that we never have to estimate densities $P(X_a)$ or $P(X_b)$ explicitly; it suffices to optimize a scalar-valued dual function $f$ over samples.

A central result in~\citet{garg2023clustering} is that, for a fixed dual function $f$, the optimal two-way partition of a region $\mathcal{R}$ that maximizes the empirical KL objective is \emph{contiguous} in the one-dimensional dual space $\{f(x): x\in\mathcal{R}\}$: there exists a cut point $c$ such that all points with $f(x)\ge c$ belong to one cluster and all points with $f(x) < c$ to the other. This reduces the combinatorial problem of optimizing labels over $\mathcal{R}$ to a one-dimensional cut search over sorted values $\{f(x)\}$, and the resulting objective is submodular, so greedy algorithms that search for cut points (or perform recursive bisection of the highest-entropy cluster) enjoy near-optimality guarantees. In JAF, we reuse this structure to define divergence-based hash predicates: a predicate $h$ at a node of the hierarchy is implemented as
\[
h(x) = \mathbf{1}\{f(x) \ge c\}
\]
for some dual function $f$ and cut point $c$ learned by maximizing (an empirical version of) the symmetric KL objective within that node. This construction yields bits that, by design, produce child regions whose empirical distributions are as distinct as possible, while remaining contiguous and easy to interpret in the one-dimensional dual space.

\begin{figure}[tp!]
  \centering
  \includegraphics[width=\linewidth]{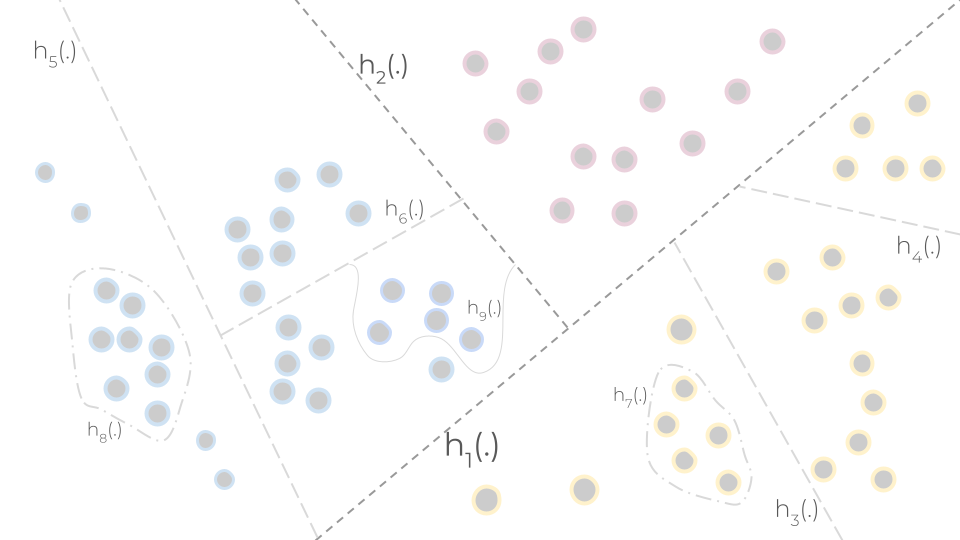}
  \caption{
  Conceptual illustration of the different types of hash predicates $h_1$–$h_9$ in JAF's LSH-based scheme.
  Each colored circle is a query--response pair $(Q_i,R_i)$, with color indicating coarse side information or label $Y$ (e.g., vulnerability / asset category).
  $h_1$ (large diagonal dashed line) is a global, label-aware predicate that makes the first coarse split, separating the yellow points from the blue and red ones.
  Within the region selected by $h_1$, $h_2$ (as a local hash function) separates red points from the blue points.
  %
%
  $h_3$, $h_4$, $h_5$, and $h_6$ are information-theoretic clustering (ITC) predicates that act \emph{locally}.
  First, $h_3$ divides the yellow bucket into two subclusters; $h_4$ then refines one of those yellow subclusters.
  Analogously, $h_5$ splits the main blue region into two parts and $h_6$ further refines one of the resulting blue sub-buckets.
  Outside their respective buckets these bits are effectively constant, so they provide finer-grained, distribution-aware refinement without changing the global layout.
  $h_7$ and $h_8$ (closed dashed contours) are information-theoretic OOD predicates.
  In each case, a small reference subset inside the contour defines typical in-distribution behavior for that region; nearby points of the same color that deviate statistically from this reference are flagged as OOD and given a dedicated bit, carving out rare but important patterns inside otherwise homogeneous buckets.
  Finally, $h_9$ (wavy solid boundary) is a purely LLM-defined semantic predicate: the LLM proposes a binary split over the points in this region based on textual criteria, without using the ITC or OOD detectors or the underlying embedding geometry.
  Overall, only a few predicates such as $h_1$ act globally; most bits are local refinements ($h_3$–$h_6$), OOD-focused ($h_7$, $h_8$), or LLM-semantic ($h_9$).
  Thus, hashcodes correspond to paths in a hierarchical partition of the space rather than arbitrary points in a flat binary cube, providing the multi-scale, interpretable structure JAF uses to form relation-aware neighborhoods for joint judging.
  }
  \label{fig:jaf_lsh_concepts}
\end{figure}

\paragraph{OOD-driven anomaly splits as special predicates.}
The same dual-divergence machinery naturally yields ``anomaly'' predicates that isolate rare or atypical patterns inside an otherwise coherent region, building on our dual-divergence OOD framework~\citep{garg2023ood}. Given a region $\mathcal{R}$ and a small reference subset $X_{\mathrm{ref}} \subset \mathcal{R}$ representing typical or in-distribution behavior (for example, triage outcomes that are historically reliable for a specific tenant and asset class), we treat the remaining points $X = \mathcal{R} \setminus X_{\mathrm{ref}}$ as candidates and estimate the KL divergence of $X$ with respect to $X_{\mathrm{ref}}$ in dual form:
\[
\hat{D}(X \,\|\, X_{\mathrm{ref}})
=
\max_{f \in \mathcal{H}}
\left[
\frac{1}{|X|}
\sum_{x \in X} f(x)
-
\log\left(
\frac{1}{|X_{\mathrm{ref}}|}
\sum_{x' \in X_{\mathrm{ref}}} e^{f(x')}
\right)
\right].
\]
Here $f:\mathcal{X}\to\mathbb{R}$ is a dual function drawn from a fixed function class $\mathcal{H}$ (e.g., neural networks), and the expectations in the Donsker--Varadhan dual are replaced by empirical averages. In the optimized dual space $f^*(x)$, reference points $x' \in X_{\mathrm{ref}}$ concentrate near lower values, while atypical candidates $x \in X$ attain higher values. A key feature of the dual formulation, established in~\citet{garg2023ood}, is that we can choose thresholds for detecting OOD points \emph{purely from the reference set}, without any tunable hyperparameters or validation on held-out OOD data.
    
In particular, two natural thresholds emerge from the theory:
\[
\tau_{\mathrm{smooth}} \;=\; \log\left(
\frac{1}{|X_{\mathrm{ref}}|}
\sum_{x' \in X_{\mathrm{ref}}} e^{f^*(x')}
\right),
\qquad
\tau_{\max} \;=\; \max_{x' \in X_{\mathrm{ref}}} f^*(x').
\]
The first, $\tau_{\mathrm{smooth}}$, is the log-mean-exp (a smooth max) of $f^*(x')$ over the reference set and arises directly from the dual objective; the second, $\tau_{\max}$, is the hard maximum. Both depend only on the reference points. Using either of these thresholds, we define an OOD predicate
\[
h_{\mathrm{ood}}(x) = \mathbf{1}\{f^*(x) > \tau\}, \quad \tau \in \{\tau_{\mathrm{smooth}}, \tau_{\max}\},
\]
which partitions $\mathcal{R}$ into a subset of points $\{x : f^*(x) \le \tau\}$ that are statistically similar to the reference behavior and a subset $\{x : f^*(x) > \tau\}$ that consists of anomalies that are statistically far from $X_{\mathrm{ref}}$. Theoretical analysis in~\citet{garg2023ood} shows that such anomaly subsets have strictly positive, quantifiable divergence from the reference: for example, when $\tau=\tau_{\mathrm{smooth}}$ and $X_{\mathrm{ood}} = \{x \in X : f^*(x) > \tau_{\mathrm{smooth}}\}$, the empirical divergence $\hat{D}(X_{\mathrm{ood}}\|X_{\mathrm{ref}})$ admits a lower bound that scales like $\log |X_{\mathrm{ref}}|$, reflecting an information-theoretic gain from having more reference samples. Computationally, once $f^*(x)$ has been evaluated for all points, detecting anomalies reduces to a single threshold comparison per point, so the decision procedure itself runs in time linear in $|\mathcal{R}|$. Empirically, this parameter-free thresholding—no tuned ``score cutoffs''—is a major strength of the dual-divergence approach, yielding highly accurate OOD and ID detections across diverse benchmarks~\citep{garg2023ood}. Within JAF, we exploit these properties to define dedicated hash predicates that carve out rare but important triage patterns as their own buckets, without requiring any task-specific tuning of thresholds.

Within JAF, these OOD-driven predicates appear naturally deeper in the hashing hierarchy, attached to buckets where subtle misconfigurations, environment-specific quirks, or rare vulnerability--asset interactions occur. From the triage perspective, this allows the system to assign dedicated bits to rare but safety-critical situations (for example, an unusual combination of security groups and shared IAM roles that expose a management plane), so that these cases can be grouped and scrutinized together in judge neighborhoods.

\paragraph{Side information and LLM-guided predicate proposals.}
Categorical side information and labels (for example, vulnerability families, asset types, deployment environments, or ground-truth risk classes) enter this construction in two ways. First, they are part of the feature representation $X$ on which predicates operate, so the divergences above are measured in a space that already encodes semantic and label-related structure. Second, they can act as weak supervision for predicate selection: among candidate splits that achieve similar divergence, we prefer those that keep coarse label groups relatively homogeneous within child regions and avoid gratuitously fragmenting rare but semantically coherent classes. Combined with the global $I(Y:C)$ objective, this ensures that codes are both label-aware and sensitive to richer, unlabeled structure.

Large language models provide an additional, complementary source of structure that integrates seamlessly with this information-theoretic framework. On one hand, an LLM can propose candidate predicates directly in natural language: given a small sample of instances from a region (for example, vulnerability--asset pairs sharing a code prefix and similar judged outcomes) and short summaries of their contexts and rationales, we prompt the LLM to suggest a simple binary distinction that meaningfully splits the sample, such as ``services with external network exposure'' vs.\ ``internal-only services,'' or ``uses credentials shared across tenants'' vs.\ ``does not.'' We then instantiate this textual predicate as a binary function $h$ and evaluate it using the same divergence and redundancy criteria as for purely learned predicates; if it yields a high-quality split, it is accepted as a hash predicate at that node.

On the other hand, an LLM can help choose reference subsets $X_{\mathrm{ref}}$ for OOD-driven predicates. Instead of specifying $h$ explicitly, the LLM is asked to identify (or synthesize) a small set of instances that are representative of ``typical'' or ``healthy'' behavior in a region (for example, correctly triaged, low-risk internal services) or of a particular sub-mode of interest (for example, cases where a particular mitigation is known to be correct). This reference set is then passed to the dual-divergence estimator, which learns a scalar function $f$ and a corresponding threshold that separates points that look like the reference from those that do not. In this hybrid scheme, the LLM operates at the semantic level, proposing meaningful candidate questions or reference behaviors, while the divergence estimator acts as a quantitative filter that enforces the information-theoretic objectives.

Across all of these mechanisms—purely divergence-based splits, OOD-driven anomaly predicates, and LLM-guided rules—the resulting predicates are incorporated into the same hierarchical partition over the cohort. Some predicates sit high in the tree and are reused across many tenants and cohorts (for example, splits by exposure or broad environment type); others sit much lower and are only meaningful within a narrow bucket or triage scenario. The dual-divergence machinery ensures that each accepted predicate (i) meaningfully separates the empirical distributions it acts on, (ii) avoids redundancy with existing predicates, and (iii) remains robust in high-dimensional, noisy settings via neural dual functions and near-optimal greedy search over one-dimensional dual spaces. This makes information-theoretic hash predicates a natural backbone for constructing the rich, triage-aware hashcodes that JAF uses to organize neighborhoods for joint judging.
        
\subsection{CoT Exploration, Iterative Self-Refinement, and Probabilistic Evaluation}
\label{sec:deepdive_jaf:cot_refine}

JAF is not only about which instances are compared, but also about \emph{how} they are reasoned about. In practice, a primary agent rarely has a single, canonical way to think through a hard triage case: it might consider several alternative chains of thought (CoTs), consult different tools, or organize evidence in different orders. The hash structure described above allows JAF to treat these reasoning paths as first-class objects, to compare them across instances, and to reuse successful reasoning styles where they work best. In this subsection we describe how JAF organizes CoTs across a cohort, how iterative self-refinement can be understood as a form of language-mediated belief propagation on the induced knowledge graph, and how repeated JAF runs yield a simple probabilistic view of acceptance decisions.

\subsubsection{Chain-of-Thought Exploration Across a Cohort}

Figure~\ref{fig:jaf_cot_refine} contrasts the usual single-instance view of CoT exploration with JAF’s cohort-level view. For a given query $Q_i$, a conventional CoT approach might sample several independent reasoning paths $R_i^{(1)},R_i^{(2)},\dots$ and then aggregate them by majority vote or confidence weighting. This is powerful but expensive: every difficult instance gets its own mini-ensemble of CoTs, with little reuse of work across instances.

In JAF, we instead embed each complete or partial reasoning trajectory into a “reasoning space’’ and apply the same hashing machinery used for instances. Concretely, for a query $Q_i$ and its $k$-th chain of thought $R_i^{(k)}$ (including tool calls and intermediate conclusions), we construct a representation $x_i^{(k)}$ and map it to a CoT code $h_{i,k}^{\mathrm{cot}}$. CoTs that exhibit similar structure—similar sequences of checks on network exposure and IAM roles, similar ways of testing whether a vulnerability is reachable, similar mitigation templates—tend to share the same code or lie within small Hamming distance. Across the cohort, these codes organize CoTs into \emph{reasoning buckets} that cut across individual queries.

This has two immediate benefits. First, when the judge reviews a specific $(Q_i,R_i)$, its prompt can include not only neighboring \emph{instances} but also neighboring \emph{reasoning patterns}: short, summarized CoTs from other issues that share similar codes. Intuitively, the judge sees “how other, similar tickets were argued’’ and can highlight where the current reasoning follows or deviates from those patterns. Second, we no longer need to generate many CoTs per hard instance from scratch: diversity of reasoning is amortized over the cohort. Successful templates discovered on some queries naturally become available as contextual exemplars for others, via shared CoT codes and hash buckets, without any explicit engineering.

In this sense, JAF encourages what one might call \emph{collective CoT exploration}: instead of thinking “ten different ways about one issue,’’ the system lets many issues “share their stories’’ and borrow reasoning styles from one another. For vulnerability triage, this mirrors how human analysts learn: once a subtle misconfiguration or a tricky interaction between ports and IAM roles has been understood in one ticket, that way of thinking tends to propagate informally to other, apparently unrelated tickets that share deeper structural similarities. Hashing and LSH play a central role here: by embedding CoTs into a vector space and then assigning them to hash buckets, JAF can rapidly retrieve diverse but structurally similar reasoning traces for any given query. Within a bucket, random sampling of CoTs provides local diversity (different instantiations of a reasoning template), while sampling across nearby buckets in Hamming space provides global diversity (different but related templates). This LSH-based organization turns what would otherwise be an unstructured set of CoTs into a searchable, multi-scale library of reasoning patterns that the judge can draw on for in-context examples.

\begin{algorithm}[t]
\caption{Iterative JAF Self-Refinement (Per-Instance Freezing)}
\label{alg:jaf_iter}
\textbf{Require:} Initial cohort $\mathcal{B}^{(0)} = \{(Q_i,R_i^{(0)})\}_{i=1}^N$, primary agent $A$,
judge LLM $J$, neighborhood constructor, per-instance minimum acceptance rounds $T_{\min}$,
global maximum iterations $T_{\max}$ (with $1 \le T_{\min} \le T_{\max}$).\\
\textbf{Ensure:} Refined cohort $\mathcal{B}^{(T^\ast)}$.
\begin{algorithmic}[1]
\STATE $T^\ast \gets T_{\max}-1$ \COMMENT{Default: use last iteration if no earlier convergence}
\FOR{$i = 1$ to $N$}
    \STATE $\text{frozen}_i \gets \textbf{false}$; $\text{acceptCount}_i \gets 0$
\ENDFOR

\FOR{$t = 0$ to $T_{\max}-1$}
    \STATE $\mathcal{B}^{(t)} \gets \{(Q_i,R_i^{(t)})\}_{i=1}^N$

    \FOR{$i = 1$ to $N$}
        \STATE Construct $S_i^{(t)}$ using kNN or LSH on $\mathcal{B}^{(t)}$
        \STATE \COMMENT{Frozen or active instances may appear as neighbors}
    \ENDFOR

    \FOR{$i = 1$ to $N$}
        \IF{not $\text{frozen}_i$}
            \STATE $(J_i^{l,t}, J_i^{c,t}) \gets J\bigl((Q_i,R_i^{(t)}), S_i^{(t)}\bigr)$
        \ENDIF
    \ENDFOR

    \FOR{$i = 1$ to $N$}
        \IF{$\text{frozen}_i$}
            \STATE $R_i^{(t+1)} \gets R_i^{(t)}$ \COMMENT{Carry forward unchanged}
        \ELSE
            \IF{$J_i^{l,t}$ indicates refinement}
                \STATE $R_i^{(t+1)} \gets A(Q_i, J_i^{c,t}, \text{history}_i)$
                \STATE $\text{acceptCount}_i \gets 0$
            \ELSE
                \STATE $R_i^{(t+1)} \gets R_i^{(t)}$ \COMMENT{Accepted this round}
                \STATE $\text{acceptCount}_i \gets \text{acceptCount}_i + 1$
                \IF{$\text{acceptCount}_i \ge T_{\min}$}
                    \STATE $\text{frozen}_i \gets \textbf{true}$
                \ENDIF
            \ENDIF
        \ENDIF
    \ENDFOR

    \IF{all $\text{frozen}_i$ are \textbf{true}}
        \STATE $T^\ast \gets t$
        \STATE \textbf{break} \COMMENT{All instances have been stably accepted}
    \ENDIF
\ENDFOR
\end{algorithmic}
\end{algorithm}
        
\subsubsection{Iterative Self-Refinement as Belief Propagation}

The same hash structure also supports iterative self-refinement at the cohort level. Figure~\ref{fig:jaf_bp} and Algorithm~\ref{alg:jaf_iter} illustrate the process. At a high level, JAF runs in rounds:-
    
\textbf{1).} The primary agent produces initial responses $R_i^{(0)}$ for all queries $Q_i$ in a cohort.

\textbf{2).} In round $t$, for each instance $i$ that is still under refinement, the judge reviews $(Q_i,R_i^{(t)})$ in the context of a small, hash-guided neighborhood $S_i^{(t)}$ of other query--response pairs, producing a binary label $J_i^{l,t}$ (accept vs.\ refine) and a critique $J_i^{c,t}$.

\textbf{3).} The primary agent then has the option to refine each active response using that critique (and any internal state or history it maintains), yielding updated responses $R_i^{(t+1)}$ for active instances; responses for already ``frozen'' instances are simply carried forward unchanged.

\textbf{4).} Neighborhoods are recomputed based on the updated cohort, and the process repeats until either (i) every instance has been accepted for at least a fixed number of rounds $T_{\min}$ (so that each has been judged under multiple, diverse local neighborhoods), or (ii) a global iteration cap $T_{\max}$ is reached. Crucially, once an instance has satisfied the $T_{\min}$-round acceptance criterion, it is \emph{excluded} from further judge calls and refinements, even as the loop continues for the remaining instances.
    
Formally, let $\mathcal{B}^{(t)} = \{(Q_i,R_i^{(t)})\}_{i=1}^N$ denote the cohort at iteration $t$, with $R_i^{(0)}$ produced without judge feedback. At each iteration, we construct neighborhoods $S_i^{(t)}$ (using kNN or LSH) on $\mathcal{B}^{(t)}$, but we only query the judge and update responses for instances that are still marked as active (i.e., not yet frozen). This per-instance freezing allows some responses to stop changing once they have been stably accepted, while still letting corrections and beliefs propagate through the rest of the cohort. As the refinement loop progresses, the set of active instances shrinks, reducing compute, but the full original cohort remains available as a source of neighbors: both active and frozen instances participate in neighborhood construction. In fact, frozen instances---having survived several rounds of scrutiny---often provide higher-quality, less noisy in-context exemplars for the judge, so keeping them in the neighborhood pool is particularly valuable for guiding the refinement of more difficult cases.
    
Because neighborhoods are recomputed at each iteration, and because LSH-based neighborhoods are randomized within buckets and across nearby buckets, the effective adjacency structure $G^{\mathrm{KG},(t)}$ of the cohort-level knowledge graph changes over time. Recomputing neighborhoods does not necessarily require retraining or modifying the underlying hash functions or neural nets at every iteration: in the simplest deployment, we hold the hash predicates fixed over several rounds, recompute hashcodes for the updated responses $R_i^{(t)}$, and then re-sample neighbors stochastically from the resulting buckets. When the hashing machinery is lightweight, one may choose to occasionally update the hash functions themselves after a few iterations, but this is optional; the key requirement is that each round sees fresh, partially overlapping neighborhoods due to both updated responses and randomized sampling within buckets. An insight discovered for instance $j$ at iteration $t$ (for example, that a claimed ``internal-only'' service is in fact internet-exposed) can influence its neighbors in $S_j^{(t)}$ immediately, and then influence their neighbors at iteration $t+1$ as those instances appear in new neighborhoods. Over a few iterations, this leads to a form of multi-hop propagation of corrections and calibration, similar in spirit to belief propagation on knowledge graphs (Pearl, 1982; Zhu and Ghahramani, 2002), but implemented through LLM prompts and natural-language critiques over a graph-structured cohort whose edges encode actionable co-judgment rather than pure semantic similarity. Figure~\ref{fig:jaf_bp} depicts this process schematically: local updates spread along edges, gradually aligning beliefs (accept/refine decisions and rationales) across the cohort.

From a triage perspective, this is exactly what we want. Once the system has learned, for example, that a particular firewall rule or IAM pattern invalidates an earlier “safe” assessment, that realization should not remain confined to a single ticket. Instead, it should quickly ripple out to other tickets that depend on similar environmental assumptions, even if those tickets were originally judged acceptable. JAF’s iterative self-refinement loop provides such a mechanism, while keeping the primary agents and judges modular.

\subsubsection{Probabilistic Evaluation via Ensembled JAF Runs}

Finally, JAF’s stochastic components—randomized LSH neighborhoods, variability in CoT generation, and any optional randomization in neighborhood construction—make it natural to view its outputs as random variables. This suggests a simple, model-agnostic way to attach a notion of confidence or stability to each decision: treat each final decision as the outcome of an underlying randomized procedure, and to estimate its reliability by repeatedly querying that procedure under different randomized neighborhood and reasoning configurations.
    
\begin{algorithm}[t]
\caption{Probabilistic Evaluation of JAF Decisions}
\label{alg:jaf_prob_eval}
\textbf{Require:} Final cohort $\mathcal{B}^{\ast} = \{(Q_i,R_i^\ast)\}_{i=1}^N$ after self-refinement, judge LLM $J$, neighborhood constructor, number of evaluation samples $R$.\\
\textbf{Ensure:} Empirical acceptance probabilities $\{\hat{p}_i\}_{i=1}^N$.
\begin{algorithmic}[1]
\FOR{$i = 1$ to $N$}
    \STATE $\text{acceptCount}_i \gets 0$
    \FOR{$r = 1$ to $R$}
        \STATE Construct $S_i^{(r)}$ using kNN or LSH on $\mathcal{B}^{\ast}$ (with randomness in neighbor sampling).
        \STATE $J_i^{(r)} \gets J_{\text{label}}\bigl((Q_i,R_i^\ast), S_i^{(r)}\bigr)$ \COMMENT{Judge label (accept vs.\ refine) for sample $r$}
        \IF{$J_i^{(r)}$ indicates acceptance}
            \STATE $\text{acceptCount}_i \gets \text{acceptCount}_i + 1$
        \ENDIF
    \ENDFOR
    \STATE $\hat{p}_i \gets \text{acceptCount}_i / R$
\ENDFOR
\end{algorithmic}
\end{algorithm}
    
By the time we reach this evaluation phase, we assume that iterative self-refinement has already produced a final set of responses $\{R_i^\ast\}_{i=1}^N$ for a cohort $\{Q_i\}_{i=1}^N$. These responses are treated as \emph{fixed}: the primary agent is not re-invoked during probabilistic evaluation. Conceptually, a single execution of JAF to evaluate these fixed responses already involves many stochastic elements: neighborhoods are randomly sampled within and across hash buckets, and the judge may employ stochastic chain-of-thought generation. To probe the stability of decisions under this randomness, we perform, for \emph{each} instance $i$, $R$ independent judge evaluations, each with an independently sampled neighborhood (and, if enabled, independently sampled judge-side CoTs). All of these evaluations can be carried out in a single JAF ``evaluation pass'' in parallel; there is no need to rerun the entire refinement loop $R$ times. For each evaluation index $r \in \{1,\dots,R\}$, we record the binary decision $J_i^{(r)}$ produced by the judge on $(Q_i,R_i^\ast)$ under the $r$-th sampled neighborhood (and associated reasoning traces).
    
The empirical acceptance probability for instance $i$ is then
\[
\hat{p}_i \;=\; \frac{1}{R} \sum_{r=1}^R \mathbf{1}\{J_i^{(r)} = \text{accept}\}.
\]
Instances whose labels are consistently accepted across runs, with $\hat{p}_i$ close to 1 and little variation, are robustly accepted across many contexts; instances with low $\hat{p}_i$ or highly variable labels across runs are fragile and should be prioritized for human review, further automated analysis, or additional data collection. In this sense, $\hat{p}_i$ serves as an empirical \emph{consistency score} for the decision on instance $i$: high values indicate that the judgment is stable across many plausible neighborhood and reasoning configurations, while low values reveal sensitivity to the particular random choices made during JAF's operation.
    
When we refer to independently sampled CoTs in this context, we mean stochastic generation of reasoning traces used during the \emph{evaluation} of $(Q_i,R_i^\ast)$ on the judge side: for example, a judge LLM that is prompted to ``think step by step'' may produce different internal chains of thought on different samples. The primary agent’s responses $R_i^\ast$ remain fixed; we do not re-run the primary agent during this phase. Changing the random seed thus alters which neighborhoods are constructed and which judge-side reasoning paths are explored for a given instance, and taking multiple independent samples allows us to measure how robust the final binary decisions are to these variations.
    
This is closely analogous to bagging and random forests~\citep{breiman2001random}: for each instance, the $R$ evaluations can be viewed as a small “forest’’ of randomized decision trees, built from different neighborhood samples and judge-side reasoning trajectories, and the final decision aggregates across these trees via the empirical probability $\hat{p}_i$. The key difference from classical bagging is that here the randomness lies in which actionably related peers are shown to the judge and how chains of thought are sampled and reused, rather than in subsampling raw data points or features. Empirically, ensemble-style JAF evaluation yields smoother calibration curves and better separation between stable and unstable decisions in triage workloads. Across cohorts, one can monitor the distribution of $\hat{p}_i$ values as a coarse-grained measure of system-level consistency: a well-calibrated, reliable JAF deployment will exhibit a large fraction of instances with high $\hat{p}_i$ and a relatively small tail of low-consistency, high-uncertainty cases that warrant special attention.
    
\subsubsection*{}
Overall, JAF can thus be viewed as a compositional mechanism for cohort-aware judgment: hash-induced neighborhoods define local, actionably related contexts; the judge LLM transforms focal responses based on those contexts; iterative self-refinement and randomized neighborhood sampling propagate corrections and align beliefs across the induced knowledge graph; and multiple randomized evaluations per instance of the fixed, refined responses $\{R_i^\ast\}$ provide probabilistic acceptance estimates based purely on binary judgments.
    
\subsection{RL-Guided Exploration: High-Level Outlook}
\label{sec:deepdive_jaf:rl}

The hash structure and CoT organization described above also suggest a natural role for reinforcement learning (RL), but a full treatment of RL for JAF is beyond the scope of this paper and will be the focus of separate work. Here we briefly sketch how RL could be layered on top of JAF’s discrete, hash-organized reasoning space as a mechanism for \emph{adaptive test-time compute allocation} and \emph{guided exploration of reasoning patterns}.

Recent RL-for-reasoning approaches typically operate at the level of a \emph{single query}: for a hard input, many chains of thought are sampled, a reward model scores them against reference answers, and the policy is updated to favor high-reward trajectories. This is effective when reference answers are easily available and when a moderate number of trajectories is likely to contain at least one high-quality reasoning path. In complex triage settings, however, a truly satisfactory CoT may require subtle integration of hundreds of implicit signals (assets, environments, mitigations, dependencies), and even large numbers of randomly sampled trajectories can easily miss such paths altogether. In contrast, JAF’s hash-based organization of CoTs and instances provides a structured, cohort-level space in which RL can operate: policies can reason over buckets of related issues and reasoning styles, rather than over isolated, per-query samples.

At a coarse level, each hash bucket groups together instances that the learned predicates deem actionably related for joint judging. Over time, JAF can accumulate simple statistics per bucket: how many instances have been seen, how often judges request refinement versus acceptance, how frequently human overrides occur, and how diverse the observed chains of thought and tool usage have been. An RL policy can then treat these bucket-level summaries as states and learn how to allocate compute: which buckets deserve extra neighborhood size or more aggressive CoT exploration (because they are historically error-prone or under-explored), and which buckets can be handled cheaply (because past decisions in those regions have been stable and accurate).

Similarly, the same ideas can be applied at the level of CoT buckets. Once CoTs themselves are hashed and grouped by similarity, a policy can learn where to focus additional test-time CoT generation and where to stop early: for some regions and query types, a single, canonical reasoning template may suffice; for others, multiple diverse templates may be needed to reliably surface edge cases. Crucially, all of this can be done in a \emph{hash-aware} way, so that global test-time compute is distributed unevenly across the space of instances and reasoning paths, rather than being spent uniformly. Instead of blindly generating many trajectories for a single hard query, an RL policy operating in JAF’s hash space can choose to: (i) re-use and adapt reasoning templates that have worked well in similar buckets, (ii) deliberately explore under-populated CoT buckets to discover new templates, or (iii) avoid repeatedly sampling reasoning styles that historical evidence shows to be systematically misleading. This provides a more sample-efficient and semantically grounded alternative to per-instance RL over CoTs.

We emphasize that these RL components are not required for JAF to function: the core framework—hash-based neighborhoods, joint judging, CoT sharing, iterative self-refinement, and ensemble-style probabilistic evaluation—can be implemented entirely through in-context prompting and lightweight auxiliary models. RL becomes attractive when one wishes to push JAF toward highly optimized, adaptive regimes, where every unit of test-time compute is strategically allocated to the buckets and reasoning templates that benefit from it most. We leave the design, analysis, and empirical evaluation of such hash-aware RL policies for JAF to future, more specialized work.

\subsubsection{Interpretable Hashcode Summaries for RL and Operators}

For RL policies and human operators alike, it is useful to have both numeric and semantic views of JAF’s hashcodes. Numerically, each bucket can be associated with statistics such as its size, the empirical acceptance rate, the frequency of human overrides, local measures of CoT diversity, and simple error indicators (e.g., disagreement between judge runs). These statistics form a compact state representation that an RL policy can consume directly.

Semantically, we can also summarize hash buckets in natural language. Given a small sample of instances from a bucket—along with their metadata (asset type, environment, vulnerability family, IAM patterns, etc.) and perhaps short rationales—we can prompt an LLM to produce a concise description such as ``internet-exposed web services on production assets with permissive security groups'' or ``internal-only analytics jobs on non-critical data, using a legacy client library.'' Alternatively, in settings with rich structured metadata, we can assemble such summaries via simple templates over categorical features.

These interpretable summaries serve two purposes. First, they provide human operators with an at-a-glance understanding of what each bucket represents, making JAF’s behavior and any downstream RL policy behavior easier to inspect, debug, and align with domain expectations. Second, they can be fed back into an RL policy as text—for example, by conditioning a policy LLM on both the numeric statistics and a textual description of the bucket—so that control decisions are informed not only by raw counts and rates, but also by a high-level semantic picture of the region of hash space being acted upon. In this way, hashcodes become more than opaque binary strings: they act as pointers into a semantically meaningful partition of the problem space, which both RL policies and humans can reason about.

Beyond RL, these summaries can also support other operational tasks. For example, they can be used to define dashboards that track how many issues fall into particular ``risk neighborhoods'' over time, to drive rule-based alerting (e.g., if a bucket corresponding to ``external-facing services with shared IAM roles'' suddenly grows), or to help analysts rapidly understand why certain issues are being grouped together and treated similarly by JAF.

\subsection{Connections to Supervised Fine-Tuning (SFT)}
    
An alternative way to exploit cross-instance structure is through supervised fine-tuning (SFT): instead of (or in addition to) using a judge to refine responses in context, one can periodically train or adapt the underlying model on the (possibly noisy) query--response pairs it has produced. In standard SFT, each example consists of a concatenated input--output pair $(Q,R)$, and the model is trained via a cross-entropy objective to increase the likelihood of $R$ given $Q$. When applied repeatedly over time as more data accumulates, such training acts as a form of global parameter adaptation: gradients from many instances jointly adjust the model’s weights, implicitly denoising and aligning its behavior across related inputs. Conceptually, this has a similar goal to our hash-guided, judge-mediated self-refinement at inference time—leveraging signals from many instances to produce more consistent behavior—but SFT realizes this via weight updates, whereas our approach realizes it via repeated in-context judgment and hash-based retrieval, without changing parameters.

One might imagine replacing or augmenting the judge with a training loop: (i) run the primary agent on the current workload to produce responses; (ii) treat these (possibly noisy) responses as targets and fine-tune the primary agent or the judge model; (iii) rerun inference with the updated model; and repeat. In principle, both the primary agent and the judge could be refined in this way, with training losses defined on final answers, intermediate rationales, or judge labels. In practice, however, iterating tightly between large-scale training and inference on evolving responses is often impractical. Early responses from a primary agent are noisy, and they improve gradually as self-refinement proceeds; naively feeding these intermediate outputs back into SFT risks reinforcing incorrect patterns. Moreover, adapting or SFT-ing a model and redeploying it remains orders of magnitude more expensive than running a judge in context, especially in latency-sensitive or multi-tenant environments. These challenges apply whether one is fine-tuning the primary agent, the judge, or both: all require high-quality targets and non-trivial retraining infrastructure.

By contrast, our inference-time pipeline keeps the primary agent’s parameters fixed, uses the judge and hash-based neighborhoods to gradually denoise and align responses across the cohort, and only treats the final refined responses $\{R_i^\ast\}$ as candidates for any downstream training. From this perspective, inference-time self-refinement acts as a filter or pre-conditioner for SFT: it produces higher-quality, more self-consistent pseudo-labels that could safely be used for occasional, offline fine-tuning, while avoiding a tightly coupled back-and-forth loop between generation and training at runtime. From a theoretical standpoint it remains an open question in which regimes parameter-space adaptation (via SFT) and in-context, hash-guided refinement (via our JAF pipeline) are preferable; we view systematic comparison of these approaches as an important direction for future work, especially when training the judge versus the primary agent is considered.

Even when SFT is available and affordable, the LSH structure over instances and hash buckets can inform \emph{curriculum design} for training. Rather than sampling fine-tuning batches uniformly at random from a large pool of historical interactions, one can use hashcodes to: (i) stratify samples by bucket, ensuring that each batch covers a diverse set of environments and reasoning patterns; (ii) over-sample under-represented or historically error-prone buckets to focus learning pressure where it is most needed; or (iii) gradually expand the set of buckets presented during training, mimicking a curriculum that moves from simple, homogeneous regions of hash space to more heterogeneous, challenging ones. In this view, LSH not only powers our inference-time mechanisms (neighborhood selection, CoT sharing, joint judging) but also provides a natural indexing scheme for organizing, scheduling, and prioritizing data in any SFT pipeline that operates over the same triage workloads.

More broadly, inference-time hashing and SFT occupy complementary points in a design space. Our hash-based, judge-mediated pipeline emphasizes on-the-fly, instance- and cohort-specific adaptation via in-context reasoning and retrieval, without changing model parameters. SFT emphasizes longer-term, parameter-space adaptation, potentially improving performance on future workloads at the cost of heavier computation and deployment overhead. The same considerations apply whether one fine-tunes the primary agent, the judge, or both: SFT can consolidate successful patterns discovered during inference into the underlying models, but need not—and likely should not—replace inference-time refinement. We expect practical systems to combine elements of both: hash- and judge-based mechanisms to rapidly adapt and debug behavior on a live cohort, and occasional SFT updates—possibly guided by hash-based curricula—to consolidate those adaptations into the underlying models once responses have been sufficiently denoised and validated.

\section{Experimental Evaluation of JAF for Triage of Cloud Misconfigs}
\label{sec:experiments}

In the earlier sections we focused on vulnerability triage, but in realistic cloud environments vulnerabilities and misconfigurations almost always co-exist, and their most serious impacts arise from their combination: a seemingly medium vulnerability on a package may become critical when paired with overly permissive capabilities or exposed management ports, while a severe-looking CVE can be neutralized in practice if the surrounding configuration is sufficiently hardened. Cloud misconfigurations are therefore a natural testbed for JAF: they are frequent, highly context-dependent, and their impact depends on subtle interactions among software requirements, deployment choices, and mitigations-in-place. Unlike an individual CVE on a single package, a misconfiguration rarely stands alone: multiple misconfigs on the same asset (or tightly coupled subnet) are often introduced by a common cause and are frequently mitigated or exacerbated together. In our experiments this asset-level grouping is part of the task definition itself, independently of JAF: each asset together with its full set of misconfigs is treated as a single triage issue that the primary agent must analyze jointly, while JAF operates \emph{across} these issues, allowing the judge to review the misconfigs on one asset in the context of how misconfigs on other, similar assets have been analyzed.

\subsection{Cloud Misconfiguration Triage: Problem and Data}
\label{sec:experiments:taskdata}
        
In practice, triaging misconfigurations is itself a full-fledged security problem that naturally calls for an agentic framework. A realistic misconfig triage pipeline would involve distinct but cooperating agents: one to perform software–misconfig analysis (which software, if any, genuinely requires which privileged settings), another to discover mitigations-in-place or reachable from the same or adjacent assets, others to reason about network and IAM exposure (including which assets can reach which others, and under what identities), and finally one or more agents to synthesize these signals into an overall risk assessment. This decomposition mirrors how human security teams work: specialists reason about software requirements, platform hardening, and exposure, and then jointly decide which misconfigs must be fixed and with what urgency. It also highlights why an approach like JAF is attractive here: decisions about one misconfigured asset are rarely independent of how similar issues have been analyzed elsewhere in the same environment.

In this paper we focus our quantitative evaluation on the software–misconfig analysis component, because it is both central and independently challenging, and it allows a clean comparison between traditional judge-based self-refinement and JAF-based refinement. Other agentic components—such as agents dedicated to mitigation discovery, cross-asset network and IAM reasoning, and joint vulnerability–misconfig triage—are part of our broader system and exhibit qualitatively similar benefits from JAF, but their detailed empirical analysis is more intricate (since each agent’s effectiveness depends on the outputs of upstream agents) and is therefore deferred to an extended version.

We consider a cohort of 315 assets (nodes, VMs, pods, and similar entities), each with a non-empty set of misconfigs (between 1 and 54 per asset) identified by standard scanners and configuration checks. These misconfigs span a typical range of cloud-hardening issues: overly permissive capabilities or security contexts, weak or absent authentication on management interfaces, unsafe hostPath mounts, broad IAM permissions, and similar conditions. In many cases, subsets of an asset’s misconfigs stem from a common cause—for example, a storage or backup system that has been granted extensive privileges for operational reasons, or an administrative shortcut taken during incident response that was never fully reverted.

For each asset we construct a prompt that serves as the query $Q_i$. This prompt contains:
\begin{itemize}
    \item all available details of the asset together with its role in the environment (for example, worker node, control-plane component, bastion host, storage node);
    \item key properties related to exposure and impact, including direct or indirect internet reachability, accessibility to critical resources or data on other assets, and any known or hypothesized attack chains involving this asset;
    \item the full list of misconfigs identified on the asset, together with brief scanner-style descriptions; and
    \item when relevant for understanding mitigations, a small amount of information about immediately adjacent assets in the cyber network whose configuration directly affects this asset’s risk (for example, firewalls, API gateways, or cluster control-plane nodes).
\end{itemize}
These adjacent assets are part of the \emph{single-asset} context for triage (they capture network and IAM relationships that influence the focal asset’s risk); they are not JAF neighbors in the sense of Section~\ref{sec:deepdive_jaf}, and they do not couple different triage issues at the judge layer.

Given this enriched asset view, the primary agent produces an analysis $R_i$. In the \emph{software–misconfig analysis} task, $R_i$ is required to identify which software components on or immediately around the asset genuinely require some of the reported (mis)configs for correct functioning in that environment, and, for each such component, to determine the minimal, correct, and complete subset of misconfigs that are actually required. This is non-trivial for several reasons. First, assets often host many software components, of which only a small subset may truly need elevated privileges or unusual configuration. Second, whether a given software requires a particular misconfig depends on the asset’s role and deployment mode: for example, a distributed storage system might legitimately need privileged volume mounts and broad device access on worker nodes, but not on bastion hosts where only administrative clients are installed. Third, the same software can have different configuration requirements on different assets depending on surrounding hardening and platform features. Correctly sorting “required for functionality’’ from “accidental or avoidable misconfiguration’’ therefore requires both domain knowledge of software behavior and careful reasoning about the local context.

In our production system, additional agents build on this software–misconfig layer to identify mitigations-in-place, reason about cross-asset network and IAM exposure, and produce risk assessments that reflect the joint effect of vulnerabilities and misconfigs. We expect JAF to provide similar benefits for those agents—propagating subtle insights about mitigations or exposure across cohorts—but a careful evaluation of these richer components, with their multi-stage dependencies, lies beyond the scope of this paper and will be presented in future work.

Across all of these settings, misconfigs on a given asset are triaged \emph{together} rather than independently. This mirrors operational practice: subsets of misconfigs on an asset often share causes (for example, enabling a storage system or monitoring agent), mitigations-in-place typically address multiple misconfigs at once, and realistic risk assessment depends on how the entire configuration, software set, and network position interact. It also makes the task inherently high-dimensional and multi-faceted: the agent must reason jointly over multiple misconfigs, multiple pieces of software, and multiple possible mitigating or risk-increasing factors, including reachability, IAM relationships, and asset criticality. As we will see in the remainder of Section~\ref{sec:experiments}, this structure makes misconfig triage an especially compelling testbed for JAF, because many of the most informative regularities only become visible when triage decisions are considered jointly across a cohort of related assets.

\subsection{Setup: Primary Agent, Baseline, and Vanilla JAF}
\label{sec:experiments:setup}
    
All triage reasoning in our experiments—both for the primary task agent and for the judge, with or without JAF—is performed by a single strong reasoning-oriented LLM (Llama3.3-Nemotron-Super-49B-v1.5), with different prompts implementing the distinct roles. The model first acts as the primary agent to produce an initial analysis $R_i^{(0)}$ for each asset, and is then prompted as a judge to review, critique, and potentially trigger refinement of these analyses, following the self-refinement template of Section~\ref{sec:deepdive_jaf}.
    
The primary-agent prompt asks the primary agent to read the asset context, list the software components that plausibly require some of the misconfigs, map each such component to the specific misconfigs it needs in this deployment, and justify these requirements using evidence from the prompt and standard deployment patterns. In the full-triage scenario, the same prompt additionally asks for mitigations-in-place, network and IAM exposure, and an overall risk assessment.

As a baseline, we use iterative self-refinement with an \emph{isolated} judge. For each asset, the primary agent generates $R_i^{(0)}$. The same model, prompted as a judge, then reviews only the pair $(Q_i,R_i^{(t)})$, with no additional assets in context, and either accepts the analysis or requests refinement, providing a critique in the latter case. When refinement is requested, the primary agent is re-invoked on $Q_i$ together with the critique and its own prior reasoning, yielding $R_i^{(t+1)}$. This corresponds to Algorithm~\ref{alg:jaf_iter} with empty neighborhoods. We set a small fixed number of refinement rounds and ensure that each asset is reviewed by the isolated judge that many times: if the judge has already accepted an answer in an earlier round, the same answer is simply carried forward and re-checked in subsequent rounds without further changes. At the end of these rounds we obtain a final baseline answer $R_i^{\text{Judge}}$ for each asset.
    
For the \emph{vanilla JAF} variant, the primary-agent prompt and refinement schedule are unchanged; the only modification is that the judge agent now operates in JAF mode. For each focal pair $(Q_i,R_i^{(t)})$, we construct a small neighborhood $S_i^{(t)}$ of peer query–response pairs from the same cohort and include them in the judge’s prompt, as in Figure~\ref{fig:jaf_isolated_joint}(b). To keep the setup simple and emphasize that JAF does not depend on sophisticated neighborhood machinery, we do not use embeddings or LSH here. Instead, for each $i$ we extract from $R_i^{(t)}$ the software components claimed to require misconfigs on asset $i$—an “objective’’ class signal indicating which software the current analysis considers responsible for a subset of the misconfigs—then randomly sample four \emph{positive} neighbors whose current answers mention at least one of the same components as requiring misconfigs, and four \emph{negative} neighbors chosen uniformly from the remaining assets. The judge prompt asks the judge to compare the focal answer against these peers, look for inconsistencies in which software require which misconfigs, highlight missing or spurious requirements, and reason about which overall patterns are most plausible given the cohort-level evidence. The primary agent then runs the same iterative refinement loop as in the baseline, but now each judging step uses $(Q_i,R_i^{(t)},S_i^{(t)})$ as input. After a small number of rounds we obtain a JAF-refined answer $R_i^{\text{JAF}}$.

In this paper we evaluate only this \emph{vanilla} instantiation of JAF. It already illustrates the core idea: neighborhoods are defined using simple, objective signals from the current responses (which software is said to require which misconfigs), and are re-sampled randomly at each judging step, so that information propagates across the cohort without relying on any particular neighborhood-construction representation. More sophisticated variants—using semantic embeddings, the information-theoretic LSH machinery of Section~\ref{sec:deepdive_jaf}, and richer knowledge-graph structure over assets and issues—are highly relevant in production settings and, in our internal experience, can reduce the number of refinement iterations by steering critique to the most relevant peers. A careful study of these LSH-based and graph-augmented JAF variants, including ablations and larger-scale experiments, will be presented in an extended version of this work; here we emphasize that even the simplest form of JAF is easy to implement and already has a substantial impact on agentic triage frameworks.

\subsection{Evaluation Protocol: Probabilistic Correctness}
\label{sec:experiments:evaluation}

Evaluating misconfiguration triage is not as straightforward as treating a single scalar label as ground truth. Superficially, one might hope that an overall severity class (for example, “fix now’’ vs.\ “defer’’ or Critical/High/Medium/Low) would suffice. In realistic cloud environments, however, there is extreme class imbalance: only a very small fraction of scanner findings (often well below $0.1\%$) are actually exploitable or operationally impactful, and many issues that look “high’’ in isolation turn out to be benign once environment-specific mitigations are taken into account. Getting the final label right is therefore not enough if the stated reasons are wrong or incomplete. What matters is whether the underlying reasoning is sound: whether the right software has been identified as requiring misconfigs, whether that mapping is complete, whether mitigations and exposure have been interpreted correctly, and whether small but crucial conditions that would invalidate the assessment have been noticed. This is also what makes human evaluation hard in this domain: even experts must reconstruct and check the chains of reasoning, not just glance at a label.

For evaluation we re-use the same model, under a separate evaluation prompt, as a probabilistic judge in the sense of Algorithm~\ref{alg:jaf_prob_eval}. For the software–misconfig scenario, the evaluation prompt asks the evaluator to decide, for a fixed asset and a fixed answer, whether (i) the set of software components claimed to require misconfigs is correct and reasonably exhaustive, and (ii) the mapping from each such component to specific misconfigs is factually justified and context-appropriate. It is instructed to reject answers that contain major false positives or false negatives or that rely on clearly unsupported assumptions, even if the implied overall risk might still be acceptable. For the full-triage scenario, the same evaluation prompt also considers whether mitigations-in-place, exposure analysis, and the final risk label are consistent with the evidence.

To capture the dependence of judgments on peer context, we follow Algorithm~\ref{alg:jaf_prob_eval}. For each asset $i$ and each method $m\in\{\text{Judge},\text{JAF}\}$, we fix the final answer $R_i^{(m)}$ produced by that method and then query the evaluator $R = 10$ times on $(Q_i,R_i^{(m)})$. In each trial, the evaluator sees $(Q_i,R_i^{(m)})$ along with a freshly sampled JAF-style neighborhood of peer assets built by the same simple positive/negative scheme as in Section~\ref{sec:experiments:setup}, but it does not refine the answers further. Each trial yields a binary accept or reject decision. The empirical correctness probability
\[
\hat{p}_i^{(m)} = \frac{1}{10}\sum_{r=1}^{10}
\mathbf{1}\{\text{the evaluator accepts }(Q_i,R_i^{(m)})\text{ in trial }r\}
\]
measures how consistently method $m$’s answer for asset $i$ withstands scrutiny under diverse, cohort-aware evaluation contexts. The evaluation protocol (model, prompt, neighborhood sampling) is identical for both methods; only the fixed answers $R_i^{\text{Judge}}$ and $R_i^{\text{JAF}}$ differ. In this way, we use a JAF-style evaluator not only to refine answers, but also to repeatedly stress-test fixed analyses under varying peer contexts, and we treat the stability of these judgments as a proxy for the robustness and evidential quality of the underlying reasoning, rather than merely for agreement with any single severity label.

\begin{figure}[tp!]
\centering
\subfigure[Baseline: isolated judge self-refinement after 5 iterations.]{
\includegraphics[width=0.475\linewidth]{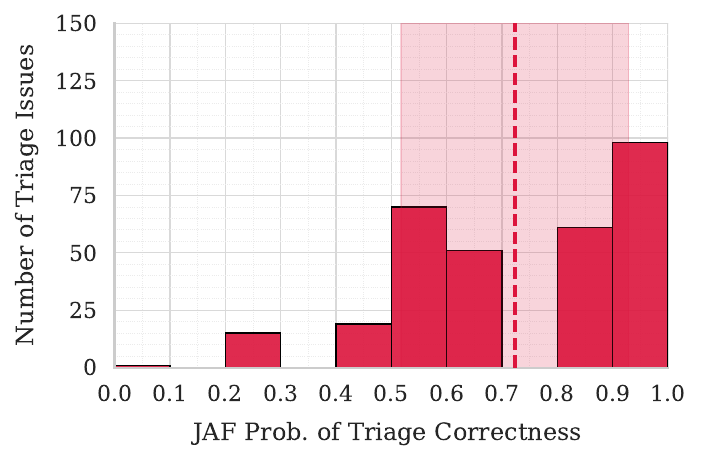}
\label{fig:jaf_eval_software_analysis_judge_5}
}
\hfill
\subfigure[JAF: cohort-aware self-refinement after 5 iterations.]{
\includegraphics[width=0.475\linewidth]{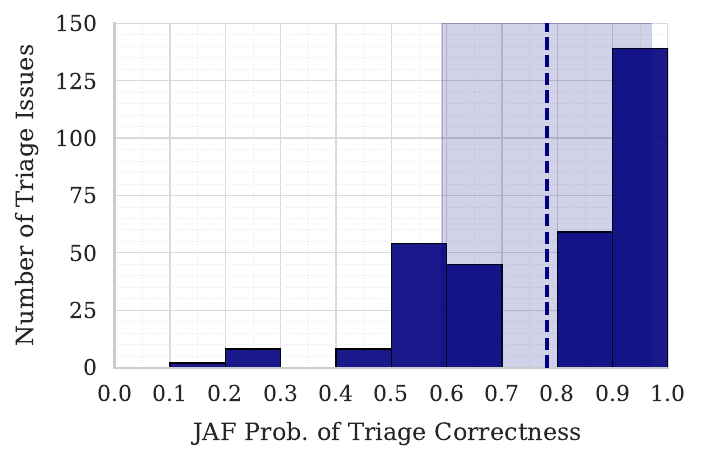}
\label{fig:jaf_eval_software_analysis_jaf_5}
}
\subfigure[Baseline: isolated judge self-refinement after 10 iterations.]{
\includegraphics[width=0.475\linewidth]{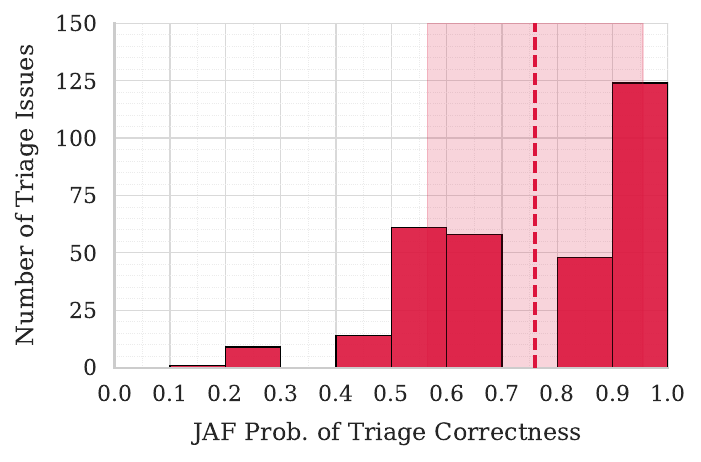}
\label{fig:jaf_eval_software_analysis_judge_10}
}
\hfill
\subfigure[JAF: cohort-aware self-refinement after 10 iterations.]{
\includegraphics[width=0.475\linewidth]{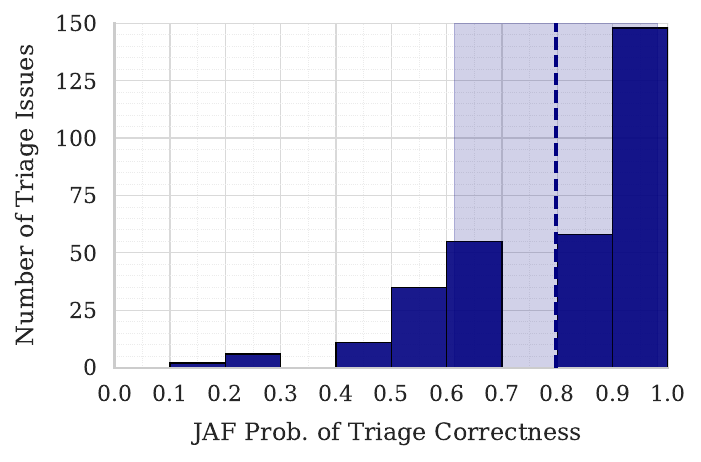}
\label{fig:jaf_eval_software_analysis_jaf_10}
}
\caption{
Distribution of empirical correctness probabilities for software–misconfig analysis on 315 assets,
shown for two refinement depths (5 and 10 iterations) and two refinement regimes (isolated judge
vs.\ JAF). For each asset and each method we fix the final primary-agent answer after the indicated
number of iterations and apply the probabilistic evaluation of Algorithm~\ref{alg:jaf_prob_eval}
with the evaluator and $R{=}10$ randomized runs. In all panels, the horizontal axis shows the
estimated probability $\hat{p}_i$ that the evaluator accepts the answer as correct under independently
sampled, cohort-aware neighborhoods, and the vertical axis shows the number of triage issues in
each bin; the dashed vertical line marks the mean $\hat{p}_i$ and the shaded band indicates one
standard deviation above and below the mean.
Panels~(a) and~(b) compare the two refinement regimes after 5 iterations. In the baseline with an
isolated judge (panel~(a)), correctness probabilities are broadly spread: many assets fall into the
ambiguous middle range ($0.3$–$0.7$), and a noticeable fraction lie below $0.5$, indicating that the
evaluator’s verdict is sensitive to which peers it sees in context. In contrast, JAF-based refinement
after the same 5 iterations (panel~(b)) already shifts the mass markedly toward higher probabilities:
more assets achieve $\hat{p}_i$ close to one, the mean increases, and the variance shrinks, reflecting
more accurate and more stable software–misconfig analyses.
Panels~(c) and~(d) show the same comparison after 10 refinement iterations. The isolated-judge
baseline (panel~(c)) does improve with extra iterations—some mass moves away from the lowest
bins and the mean $\hat{p}_i$ increases—but the distribution remains relatively broad, with a
persistent tail of low-confidence cases. By contrast, JAF after 10 iterations (panel~(d)) only modestly
sharpens what was already a concentrated distribution at 5 iterations: most assets were already
clustered near high $\hat{p}_i$ in panel~(b), and additional iterations provide diminishing returns.
Together, these trends indicate that JAF not only achieves higher correctness probabilities than the
baseline, but does so with substantially fewer refinement steps. Across all four panels, JAF also
preserves—and clarifies—the tail of genuinely hard cases: a small set of assets retains low empirical
correctness probabilities under both methods, though consistently fewer under JAF. These are
precisely the issues that should be escalated for human review or targeted investigation.}
\label{fig:jaf_eval_software_analysis}
\end{figure}

\subsection{Results and Discussion}
\label{sec:experiments:results}

Figure~\ref{fig:jaf_eval_software_analysis} summarizes how the empirical correctness probabilities $\hat{p}_i$ evolve as we increase the number of refinement iterations, for both the isolated-judge baseline and JAF-based self-refinement. The top row, panels~(a) and~(b), shows results after 5 refinement iterations; the bottom row, panels~(c) and~(d), shows results after 10 iterations. In all four panels, the horizontal axis is the estimated probability that the evaluator accepts the answer as correct under independently sampled, cohort-aware neighborhoods, and the vertical axis is the number of assets in each bin; the dashed vertical line and shaded band mark the mean and one standard deviation, respectively.

After only 5 iterations, the contrast between the two methods is already clear. In panel~\ref{fig:jaf_eval_software_analysis_judge_5}, corresponding to the isolated-judge baseline, correctness probabilities are broadly spread: many assets fall in the ambiguous middle range (roughly $0.3$–$0.7$), and a noticeable fraction lie below $0.5$. This indicates that the evaluator’s verdict on these baseline answers is sensitive to which peers it sees in context—exactly what one would expect when each asset is refined in isolation and subtle mis-assignments or omissions in the software–misconfig mapping can persist undetected. In panel~\ref{fig:jaf_eval_software_analysis_jaf_5}, JAF-based refinement after the same 5 iterations has already shifted the mass markedly toward higher probabilities: many more assets achieve $\hat{p}_i$ close to one, the mean increases, and the variance shrinks. Allowing the judge to see a small cohort of related assets when critiquing each answer leads to software–misconfig analyses that are more often judged correct and are less fragile to changes in peer context.

Increasing the number of refinement iterations from 5 to 10 further illustrates how JAF accelerates convergence. Comparing the isolated-judge baseline in panel~\ref{fig:jaf_eval_software_analysis_judge_5} to its 10-iteration counterpart in panel~\ref{fig:jaf_eval_software_analysis_judge_10}, we see that additional iterations do help: some mass moves out of the lowest-probability bins toward the middle and upper ranges, and the mean $\hat{p}_i$ increases. However, even after 10 iterations the baseline still exhibits a wide spread of correctness probabilities and a noticeable tail of low-confidence cases. By contrast, JAF’s distribution in panel~\ref{fig:jaf_eval_software_analysis_jaf_10} is only modestly sharper than in panel~\ref{fig:jaf_eval_software_analysis_jaf_5}: most assets were already concentrated near high $\hat{p}_i$ after 5 iterations, and the additional 5 iterations provide diminishing returns. Put differently, the JAF distribution after 5 iterations already resembles its 10-iteration distribution more closely than the baseline’s 10-iteration distribution resembles JAF’s 5-iteration one. This suggests that JAF is not only achieving higher final quality, but doing so with significantly fewer refinement steps.

Across all four panels, JAF also preserves—and clarifies—the tail of genuinely hard cases. A small set of assets retains low empirical correctness probabilities under both methods, though they are consistently fewer under JAF. These correspond to ambiguous or atypical situations—unusual software stacks, niche configurations, or rare mitigations—where even cohort-level comparison does not yield a stable consensus. Rather than masking such difficulty, the probabilistic evaluation makes it explicit: low $\hat{p}_i$ values naturally flag issues that should be escalated for human review or targeted data collection. In operational settings this separation between stable and fragile decisions can be as important as improvements in mean accuracy.

We observe a similar qualitative pattern in a full-triage setting (not shown), where the primary agent must in a single prompt perform software–misconfig analysis, identify mitigations-in-place, reason about network and IAM exposure, and produce a final risk score. There too, the isolated-judge baseline yields a broad spread of correctness probabilities with a pronounced low-confidence tail, whereas JAF-based self-refinement shifts the distribution toward higher correctness and tighter concentration, while preserving a small set of clearly unstable cases. Overall, these experiments show that even a very simple instantiation of JAF—without embeddings, without LSH, and with neighborhoods defined only by overlap in the software claimed to require misconfigs—can substantially improve both the accuracy and the stability of complex triage reasoning, and can do so in a way that naturally highlights the residual hard cases that most warrant human attention.

\section{Conclusion and Future Work}
\label{sec:conclusion}

JAF lifts the judge from an instance-local critic to a cohort-aware reasoner: each judgment is made not only on the focal query–response pair but also in the light of a small, relation-aware neighborhood of peer pairs drawn from the same workload. As developed in Sections~\ref{sec:intro_jaf} and~\ref{sec:deepdive_jaf}, this induces a cohort-level knowledge-graph structure and a language-mediated analogue of belief propagation and random-forest style ensembling, all within an in-context learning regime. Our preliminary experiments on misconfiguration triage show that even a naive instantiation of JAF, without embeddings or LSH, already delivers more accurate and more stable software–misconfig analyses than standard isolated self-refinement, and naturally identifies the minority of issues that remain fragile under cohort-aware scrutiny.

Several directions follow naturally. Methodologically, an immediate next step is a systematic empirical study of the LSH-based neighborhoods introduced in Section~\ref{sec:deepdive_jaf}, including ablations over hash predicates, neighborhood sizes, and refinement schedules, on larger and more heterogeneous triage workloads. On the application side, we plan to extend our evaluation beyond software–misconfig analysis to full end-to-end triage pipelines and to other domains where cohort-level judgment is central, complemented by human-in-the-loop studies that examine how analysts use JAF’s uncertainty estimates and neighborhood summaries. In the longer term, we are interested in tighter integration between JAF and learning mechanisms such as reinforcement learning over hash buckets for adaptive test-time compute allocation, and supervised fine-tuning that consolidates successful in-context adaptations into the underlying models, while preserving the modular, interpretable, and privacy-conscious structure that makes JAF attractive for deployment.
        
\bibliography{references}

\bibliographystyle{apalike}

\end{document}